\documentclass[11pt]{article}

\usepackage[final]{acl}

\usepackage{times}
\usepackage{latexsym}

\usepackage[T1]{fontenc}

\usepackage[utf8]{inputenc}

\usepackage{microtype}

\usepackage{inconsolata}

\PassOptionsToPackage{table}{xcolor}
\usepackage[most]{tcolorbox}

\usepackage[table]{xcolor}

\usepackage{graphicx}
\usepackage{amsmath}
\usepackage{algorithm}
\usepackage{algorithmic}
\usepackage{amssymb}
\usepackage{multirow}

\usepackage{CJKutf8}
\usepackage{booktabs}
\usepackage{longtable}
\usepackage{array}
\usepackage{bbm}
\usepackage{CJK}
\usepackage{xcolor}
\usepackage{spverbatim}
\usepackage{array}
\usepackage{booktabs}

\newcommand{\red}[1]{\textcolor{red}{#1}}

\title{\textsc{CLARity}: Reasoning Consistency Alone Can Teach Reinforced Experts}

\author{
    Jiuheng Lin, 
    Cong Jiang, 
    Zirui Wu, 
    Jiarui Sun, 
    Yansong Feng\thanks{Corresponding author.} \\
    $^{}$ Wangxuan Institute of Computer Technology, Peking University \\
    {\tt linjiuheng@stu.pku.edu.cn} ~~ {\tt fengyansong@pku.edu.cn} \\
}

\begin{document}
\maketitle
\begin{abstract}

Training expert LLMs in domains with scarce fine-grained annotated data is admittedly challenging, often relying on multiple-choice questions (MCQs). 
However, standard outcome-based reinforcement learning (RL) on MCQs is risky. 
While outcome-based RL may improve accuracy, it frequently compromises the reasoning process, yielding internally inconsistent rationales that diverge from the final predictions. 
Existing solutions to supervise the reasoning process, such as large-scale Process Reward Models (PRMs), are prohibitively expensive. 
To address this, we propose \textsc{CLARity}, a cost-effective RL framework that enhances reasoning quality using a small, general-purpose LLM only. \textsc{CLARity} integrates a consistency-aware reward mechanism with a 2-stage refine-then-monitor training pipeline to enhance reasoning consistency, and a dynamic data reformulation strategy to better exploit annotated data available. 
Experiments demonstrate that \textsc{CLARity} can improve the consistency of responses by 16.5\% over standard outcome-based RL, 
and bring an improvement of 7.5\% in final accuracy. 
Human evaluations further confirm substantial gains in factual correctness and reasoning coherence, leading to more trustworthy model outputs.
Thus, \textsc{CLARity} offers a generalizable solution that enables smaller models to effectively guide expert LLM training by monitoring reasoning consistency.\footnote{Our code is open sourced at: \url{https://github.com/Infinite-set/CLARity}}

\end{abstract}

\section{Introduction}

Reinforcement learning (RL) has recently become as a popular paradigm for enhancing the reasoning ability of large language models (LLMs)~\cite{guo2025deepseek,team2025kimi, qwq32b}, 
yielding significant improvements in math~\cite{fengReToolReinforcementLearning2025,yang2025reasonfluxhierarchicalllmreasoning} and code~\cite{wang2025codeboostboostingcodellms,fan2025posteriorgrporewardingreasoningprocesses}. 
However, in other domains such as law and medicine, 
high-quality training data is often scarce. 
Available resources are typically limited to materials like professional qualification exams, where the predominant format is the multiple-choice question (MCQ).

\begin{figure}[t]
\centering
\includegraphics[scale=0.635]{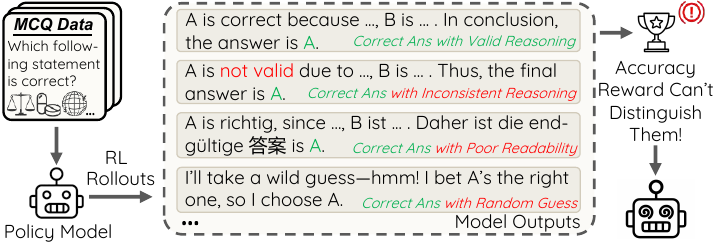}
\caption{Illustration of risks in MCQ RL: rewarding only answer correctness neglects reasoning supervision, which may weaken reasoning quality during training.}
\label{fig:introduction}
\end{figure}

Training using MCQs is usually effective, as their deterministic answers allow for clear outcome rewards, which avoids the potential instability of training on open-ended tasks where a gold standard answer is sometimes ambiguous~\cite{5team2025glm45agenticreasoningcoding,he2025selectknowinternalexternalknowledge, weng2024rewardhack}. 
However, it also introduces critical risks: 
focusing solely on the final choice accuracy provides no guarantee for a reliable reasoning process. 
As shown in Figure~\ref{fig:introduction}, the model may learn to find correct answers through flawed heuristics or random guessing rather than acquiring robust reasoning abilities. 

We conduct a pilot study by training an LLM on a judicial examination MCQ dataset using standard outcome-based RL. 
As shown in Figure~\ref{fig:method:pilotstudy}, 
while the final-answer accuracy significantly increases, the overall response quality progressively degrades, manifesting as a sharp rise in \textit{reasoning inconsistency}---a phenomenon where the model's intermediate analytical steps explicitly contradict its final chosen answer. 
We observe that the proportion of such inconsistent responses increases from 7\% to 31\%. 
This indicates that standard RL on MCQs may corrupt the reasoning process, resulting in models that are statistically accurate yet driven by totally unreliable rationales.

Solving MCQ response quality issues is nontrivial. 
In domains like mathematics or code, the reasoning correctness can often be validated programmatically by external verifiers such as the Python or Lean~\cite{lei2025reasoningtableexploringreinforcementlearning,zhao2025agentict2srobusttexttosparqlagenticcollaborative}. 
For domains lacking such verifiers, current studies often rely on Process Reward Models (PRMs) to monitor the reasoning process. 
These approach, however, typically require either the use of large commercial models like GPT-4o as the PRM, or the fine-tuning of smaller LLMs on substantial high-quality, expert-annotated data~\cite{chen2024huatuogpto1medicalcomplexreasoning, su2025crossingrewardbridgeexpanding}. 
Both methods are resource-intensive, demanding significant budget or large expert-labeled corpora.

To overcome these challenges, we introduce \textbf{C}onsistency-aware \textbf{L}earning with Data-\textbf{A}ugmented 2-Stage \textbf{R}e\textbf{i}nforcemen\textbf{t} Strateg\textbf{y} (\textsc{CLARity}), 
an MCQ RL framework designed to enhance response quality using only a small, general-purpose LLM, without requiring any further fine-tuning with domain-specific annotated data. 

\textsc{CLARity} employs a consistency reward that assesses reasoning consistency within the response, penalizing incoherent outputs. 
The reward is integrated into RL via a two-stage refine-then-monitor pipeline: Stage-1 refines the model's output to promote transparent, option-wise reasoning structure; 
Stage-2 relaxes format constraints and targets deeper reasoning, enhancing reasoning flexibility while preventing reward hacking. 
Additionally, to better exploit scarce data, \textsc{CLARity} also apply a dynamic data reformulation strategy to improve data efficiency and overall training outcomes.

Our experiments validate the effectiveness of \textsc{CLARity}, which achieves improvements of 16.5\% in response consistency and 7.5\% in reliable reasoning accuracy over standard RL baselines. 
Human evaluations further confirm that by focusing solely on the simple signal of logical consistency, \textsc{CLARity} produces models with \textbf{reliable reasoning abilities} that are not only more coherent but also exhibit greater factual correctness and readability. 
Moreover, by eliminating the need for large-scale teacher LLMs or expert-annotated datasets, 
\textsc{CLARity} offers a cost-effective and generalizable solution across domains, 
enabling smaller, general-purpose LLMs to effectively guide the training of expert models to achieve both higher accuracy and superior reasoning quality.

Our contributions are summarized as follows:
(1) We identify MCQ RL can yield superficially accurate yet unreliable, inconsistent reasoning. 
(2) We propose \textsc{CLARity}, a novel MCQ RL framework that integrates consistency-aware mechanism with a refine-then-monitor pipeline to enhance reasoning consistency and a dynamic data reformulation strategy to maximize data utility. 
(3) We show \textsc{CLARity} improves both accuracy and reliability, enabling smaller models to guide experts without requiring expert-level domain knowledge.

\section{Pilot Study: Response Quality}

we conduct a pilot study to investigate the response quality dynamics during outcome-based RL training. 
We utilize the JEC-QA case-analysis dataset~\cite{zhong2020jec}, a challenging legal benchmark that requires complex multi-step reasoning to solve multiple-choice questions (MCQs). 
We assess response quality using DeepSeek-V3 as an automated judge by evaluating \textit{reasoning consistency}---the logical alignment between a model's internal analytical trace and its final predicted option (see Appendix~\ref{app:implementation:prompt} for detailed evaluation criteria). 

\begin{figure}[t]
\centering
\includegraphics[scale=0.263]{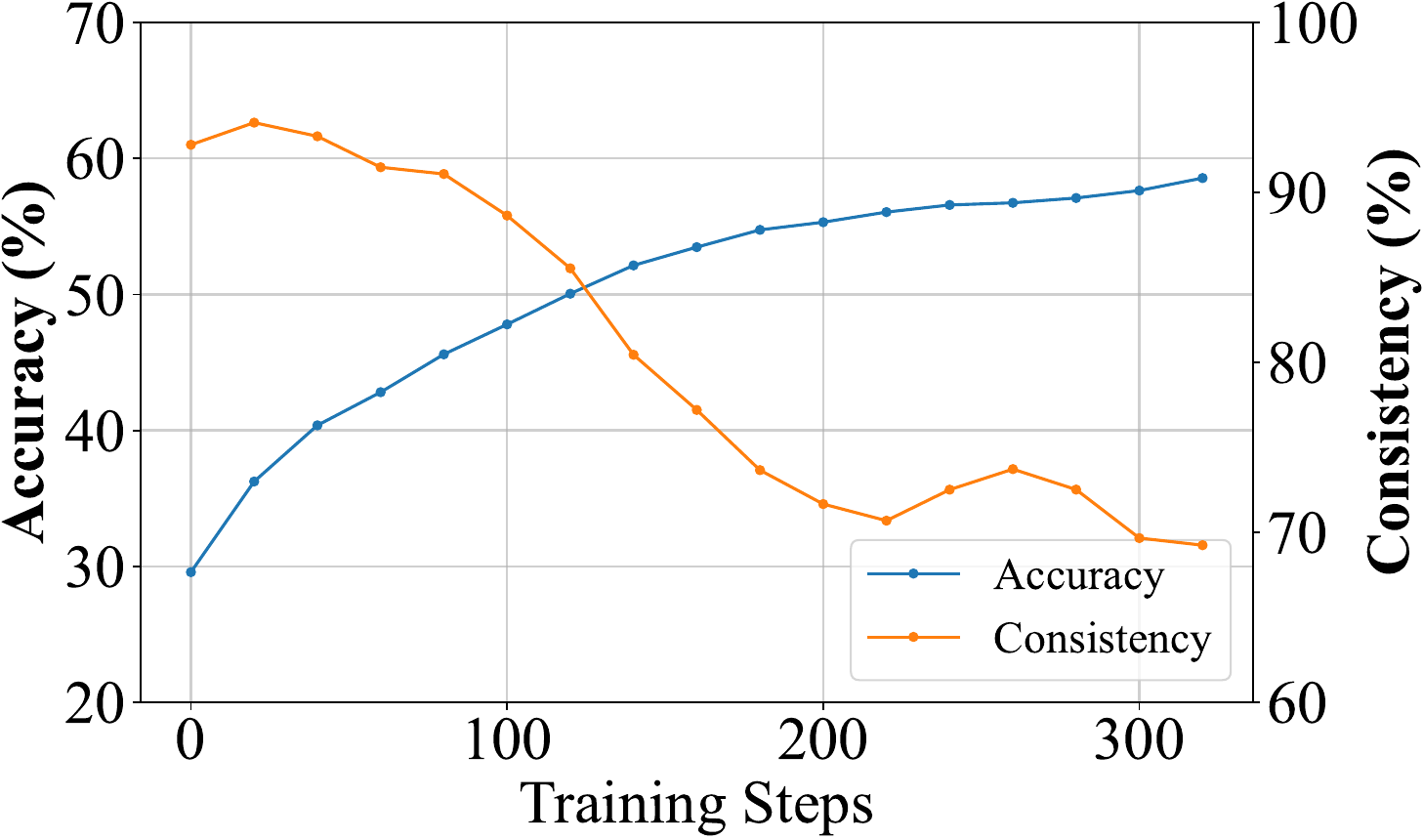}
\caption{Response quality dynamics under GRPO training. The logical consistency declines over time.}
\label{fig:method:pilotstudy}
\end{figure}

As shown in Figure~\ref{fig:method:pilotstudy}, 
while the accuracy on final answers increases, 
we observe the overall response quality progressively degrades during training. 
These findings show that MCQ RL fails to enhance reasoning, producing models that perform reasoning unreliably and inconsistently. 
Given limited MCQ data resources without expert-annotation corpora or large PRMs, 
we explore efficient strategies to address these challenges.

\begin{figure*}[t]
\centering
\includegraphics[scale=0.56]{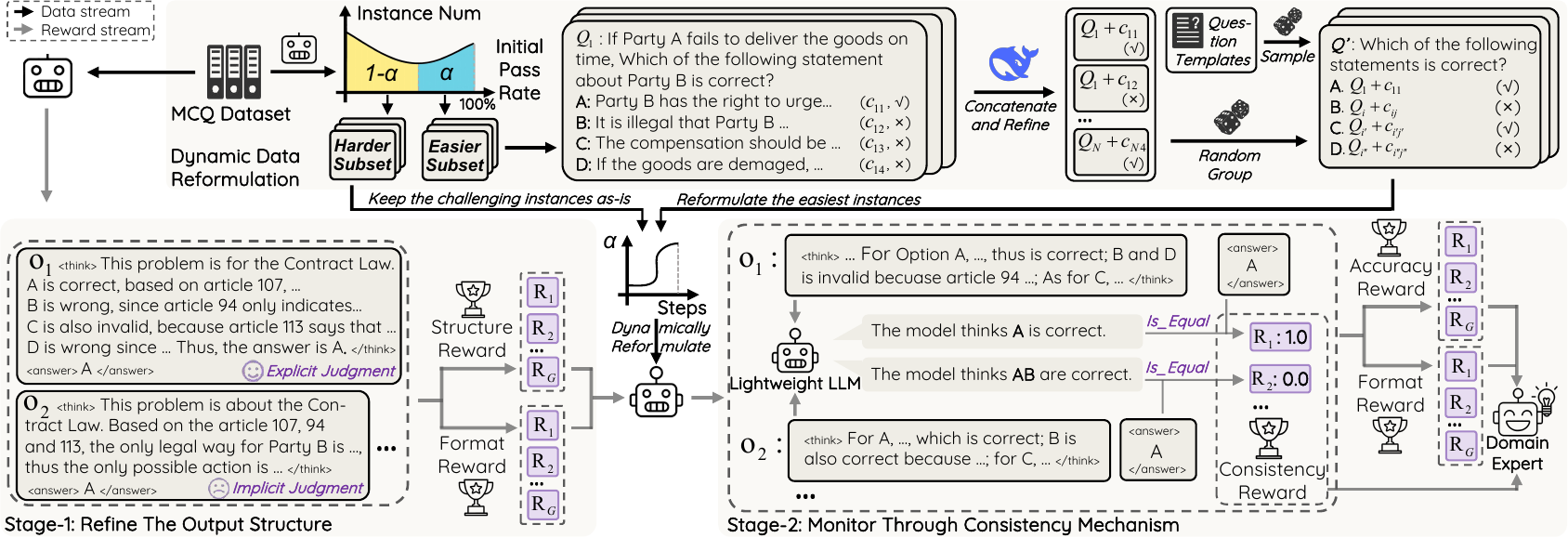}
\caption{Overview of \textsc{CLARity}, 
It combines a consistency mechanism for detecting inconsistencies,
a refine-then-monitor training pipeline for improving reasoning quality,
and a dynamic data reformulation for maximizing data utility.}
\label{fig:methodology}
\end{figure*}

\section{Methodology}

We propose \textbf{C}onsistency-aware \textbf{L}earning with Data-\textbf{A}ugmented 2-Stage \textbf{R}e\textbf{i}nforcemen\textbf{t} Strateg\textbf{y} (CLARity) to address response quality issues in MCQ RL. 
As shown in Figure~\ref{fig:methodology}, our approach integrates a consistency-aware learning mechanism into the data-augmented refine-then-monitor reinforcement pipeline. 
\textsc{CLARity} enables smaller, general-purpose LLMs to guide expert model training, improving both reasoning quality and accuracy. See the complete algorithm in Algorithm~\ref{alg:data}.

\subsection{Consistency-Aware Learning Mechanism}

Our pilot study shows that outcome-based RL on MCQs fails to improve reasoning, 
often producing responses with severe inconsistencies. 
Moreover, fully supervising the reasoning process requires either large commercial LLMs or costly expert-annotated corpora, which are often unavailable. 
Here, we explore another strategy to address the quality issue on MCQ RL, 
leveraging the consistency of model's judgments for each candidate option in its thinking trajectory as a reward signal. 
This is because a high-quality response must not only provide the correct final answer but also accurately judge the validity of each option, making consistency a useful proxy for reasoning reliability.

Specifically, our consistency-aware reward mechanism begins by separating the model's responses into two parts: the reasoning trajectory and the final answer. 
We then employ a separate, small general-purpose LLM as the consistency reward model, tasked with identifying the believed-correct options that the model endorses through its reasoning. 
A penalty is assigned if either the reward model fails to identify clear option judgments from the reasoning or the believed-correct options in the reasoning do not match the model's final answer. 
By penalizing such failures, our mechanism guides the model toward more coherent and reliable reasoning.

A key advantage of this mechanism is its minimal requirement for domain knowledge.
The reward model only needs to comprehend basic correctness judgments (e.g., "\textit{Option A is correct,}" "\textit{B is invalid because ...}") within the response, 
which relies on fundamental semantic understanding, a capability already possessed by most general-purpose LLMs. 
Thus, expert-level models are unnecessary. 
Moreover, although our method focuses narrowly on logical coherence, we observe broader improvements in the holistic reasoning proficiency. See discussions in Section~\ref{analysis:human_eval}.

\subsection{Data-Augmented Two-Stage Training}
\label{method:training_pipeline}

We introduce a two-stage refine-then-monitor pipeline to effectively integrate our consistency-aware reward into RL training. 
Additionally, to better exploit limited MCQ data, 
we propose a dynamic data reformulation strategy to improve data utilization and overall training outcomes.

\paragraph{2-Stage Refine-then-Monitor Training} 

We integrate the proposed consistency reward into training process by a two-stage refine-then-monitor pipeline, 
which progressively improves consistency while preserving in-depth reasoning. 
In stage-1, we refine the model's output structure by encouraging option-wise reasoning, without considering the answer correctness. 
We apply regular expressions to identify and reward responses that analyze one option at a time, 
thus encouraging the model to explicitly state correctness judgments in reasoning.
This ensures that responses remain LLM-friendly for accurate checking by reward models, and prevents reward hacking --- if the model fails to present explicit judgments, the consistency reward model cannot reliably detect inconsistencies, which risks pushing the model toward overly superficial and simplified reasoning~\cite{chen2025reasoningmodelsdontsay}. 
Further discussion of this issue is provided in Section~\ref{analysis:two_stage}.

In stage 2, we monitor the model's responses using our proposed consistency reward model, and eliminate the requirement for response structure in stage-1 to avoid overfitting and encourage flexible reasoning. 
Additionally, we incorporate the answer reward in stage-2 to optimize answer correctness. 
We design a strict reward mechanism that provides positive feedback only when the model selects exactly all the correct options and assigns zero reward in all other cases, 
which largely reduces the possibility of shortcut-based solutions during training and compels the model to develop a deeper reasoning. 
And a fixed format reward is applied in both stages, ensuring the model generates output within <think>...</think> and <answer>...</answer> tags.

\paragraph{Dynamic data reformulation}
\label{method:data_aug}

Gradually removing easy instances and introducing harder instances during training can enhance model performance~\cite{Polaris2025, huang2025ragrladvancingretrievalaugmentedgeneration, li2025limrrlscaling}. 
Given the scarcity of high-quality training data, 
we employ a data augmentation method that does not require additional datasets. 
Instead, it refines too-easy data through deconstruction and reformulation to maximize data utility, 
curating more challenging and diverse data without the need for domain expertise.

Specifically, we first deconstruct each training instance into independent propositions by converting each candidate option into an atom statement through string concatenation of the original query statement and the option. 
The correctness of each proposition is determined by checking if the corresponding option is correct in the original problem. 
Next we refine these propositions using the LLM, 
polishing ill-formed phrasing for fluency and introducing fictional names and places to further diversify the data. 
See Appendix~\ref{app:implementation:dataset} for data examples.

During training, based on the initial pass rate, we create new questions by dynamically reformulating the easiest $\alpha$ proportion of the original instances through randomly grouping, 
with answers derived from the correctness of each proposition and statements generated using predefined templates (e.g., \textit{"Which of the following statements are correct?"}). 
The remaining $(1-\alpha)$ instances remain unchanged. 
By gradually increasing $\alpha$, we progressively eliminate overly simple instances, 
transforming them into more challenging examples that promote better learning. This strategy offers an efficient and effective way to adjust the difficulty of training data and enhance training outcomes. 

\begin{table*}[t]
\centering
\small
\setlength\tabcolsep{4pt}
\begin{tabular}{ll|cccccc|ccc|cc}
\toprule
\multicolumn{2}{c|}{\multirow{2}{*}{\textbf{Method}}} & \multicolumn{3}{c|}{\textbf{Legal Reasoning}} & \multicolumn{3}{c|}{\textbf{Medical Reasoning}} & \multicolumn{3}{c|}{\textbf{Average}}  & \multicolumn{2}{c}{\textbf{Non-MCQ}} \\
 & & Acc$^{+}$ & Cons\% & \multicolumn{1}{c|}{Acc} & Acc$^{+}$ & Cons\% & \multicolumn{1}{c|}{Acc} & Acc$^{+}$ & Cons\% & \multicolumn{1}{c|}{Acc} & Legal & Med \\
\midrule
\multirow{2}{*}{\textit{Vanilla Qwen}} & Qwen2.5$_\text{3B}$ & 17.5 & 74.3 & 19.3 & 34.7 & 77.7 & 38.2 & 26.1 & 76.0 & 28.8 & 17.7 & 17.4  \\
& Qwen2.5$_\text{7B}$ & 28.5 & 92.3 & 29.3 & 42.5 & 88.7 & 49.4 & 35.5 & 90.5 & 39.4 & 24.9 & 17.0  \\
\midrule
\multirow{2}{*}{\textit{RL Baselines}} & Standard RL & 45.0 & 69.3 & \textbf{58.5} & 57.8 & 81.3 & 63.6 & 51.4 & 75.3 & \underline{61.1} & \underline{26.6} & 19.4 \\
 & PRM RL & 42.0 & 68.2 & 54.7 & 60.2 & \textbf{96.4} & 60.8 & 51.1 & 82.3 & 57.8 & 25.5 & \underline{22.2} \\
\midrule
\multirow{4}{*}{\textit{Ours}} & Data Reformulation & 47.4 & 75.3 & \underline{56.9} & \underline{62.4} & 89.0 & \underline{65.3} & 55.1 & 82.2 & \underline{61.1} &  25.3 & \textbf{22.6} \\
 & Refine-then-Monitor & \underline{52.3} & 93.8 & 56.3 & 59.7 & 89.3 & 62.3 & \underline{56.0} & \underline{91.6} & 59.3 & 23.6 & 18.3 \\
& \textsc{CLARity}$_\text{3B}$ & 33.9 & \textbf{97.1} & 34.2  & 50.1 & 86.0 & 55.1 & 42.0 & \underline{91.6} & 44.7 & 22.4 & 19.1 \\
& \textsc{CLARity}$_\text{7B}$ & \textbf{54.4} & \underline{94.1} & 56.4 & \textbf{63.3} & \underline{89.5} & \textbf{66.2} & \textbf{58.9} & \textbf{91.8} & \textbf{61.3} & \textbf{27.2} & \textbf{22.6} \\
\midrule
 & GPT-4o & 39.5 & 96.6 & 39.9 & 83.0 & 99.2 & 86.3 & 61.3 & 97.9 & 63.1 & 27.5 & 22.2  \\
\bottomrule
\end{tabular}
\caption{Performance on validation and non-MCQ benchmarks. Best scores are \textbf{bold}, with the second \underline{underlined}. }
\label{tab:main-result}
\end{table*}

\begin{table}[t]
\centering
\small
\renewcommand{\arraystretch}{1.1}
\setlength\tabcolsep{2.5pt}
\begin{tabular}{l|cccc|c}
\toprule
\multirow{2}{*}{\textbf{Model}} & \multicolumn{2}{c|}{\textbf{Legal}} & \multicolumn{2}{c|}{\textbf{Medical}} & \textbf{Avg.} \\
 & Iter & \multicolumn{1}{c|}{Shuf} & Iter & \multicolumn{1}{c|}{Shuf} & Acc$^+_{\text{Cons\%}}$ \\
\midrule
Qwen2.5$_\text{3B}$ & $36.8_\text{66.7}$ & $4.1_\text{47.1}$ & $48.8_\text{79.9}$ & $3.3_\text{43.2}$ & $23.3_\text{59.2}$  \\
Qwen2.5$_\text{7B}$ & $54.5_\text{84.3}$ & $14.5_\text{92.0}$ & $67.8_\text{95.3}$ & $6.9_\text{71.6}$ & $35.9_\text{85.8}$  \\
\midrule
RL & $61.5_\text{87.1}$ & $14.6_\text{55.4}$ & $72.5_\text{94.7}$ & $8.5_\text{61.8}$ & $39.3_\text{74.8}$ \\
PRM RL & $63.1_\text{88.4}$ & $10.1_\text{40.4}$ & $62.3_\text{97.2}$ & $5.7_\text{83.9}$ & $35.3_\text{77.5}$ \\
\midrule
Reform & $68.9_\text{91.3}$ & $23.2_\text{78.5}$ & $\underline{77.1}_\text{95.8}$ & $12.4_\text{\underline{86.0}}$ & $45.4_\text{87.9}$ \\
Two-Stage & $\underline{70.5}_\text{\textbf{95.8}}$ & $\underline{23.7}_\text{90.2}$ & $75.1_\text{\underline{98.7}}$ & $\underline{12.7}_\text{84.5}$ & $\underline{45.5}_\text{\underline{92.3}}$ \\
\textsc{CLARity}$_\text{3B}$ & $53.6_\text{86.8}$ & $14.7_\text{\underline{97.3}}$ & $66.9_\text{95.0}$ & $8.4_\text{85.3}$ & $35.9_\text{91.1}$ \\
\textsc{CLARity}$_\text{7B}$ & $\textbf{70.5}_\text{\underline{93.3}}$ & $\textbf{28.9}_\text{\textbf{97.3}}$ & $\textbf{81.0}_\text{\textbf{99.4}}$ & $\textbf{27.5}_\text{\textbf{88.5}}$ & $\textbf{52.0}_\text{\textbf{94.6}}$ \\
\midrule
GPT-4o  & $60.0_\text{99.2}$ & $20.3_\text{93.7}$ & $88.0_\text{99.4}$ & $40.4_\text{88.4}$ & $52.2_\text{95.2}$ \\
\bottomrule
\end{tabular}
\caption{Generalizability across two content-preserving alternative formats: (1) Iterative Judgment (Iter.) and (2) Shuffled MCQ (Shuf.).}
\label{tab:main-result-generalizability}
\end{table}

\section{Experimental Setup}
\paragraph{Datasets}
\label{datasets}
We focus specifically on two domains: law and medicine, 
where large-scale professional-level datasets exist solely in multiple-choice question (MCQ) format. 
We believe the observations from these areas are generalizable to other fields.

For legal reasoning, we use the JEC-QA case-analysis dataset~\cite{zhong2020jec},
which require deep legal analysis and reasoning in complex scenarios rather than rote memorization~\cite{patterson1951case}.
The dataset contains 10,561 case-analysis MCQs, split into 80\% for training and 20\% for validation. 
For medical reasoning, we use the MedQA-USMLE dataset~\cite{jin2020diseasedoespatienthave}, 
which evaluates a physician’s ability to analyze medical concepts, apply principles and reason across multiple pieces of evidence. 
It contains 10,178 MCQs in the training set and 1,272 MCQs for validation.

\paragraph{Evaluation}
For in-domain validations, we report the ratio of \textbf{correct and consistent} instances (\textbf{Acc$^+$} for abbreviation).
This metric better reflects the model's true problem-solving ability, as it distinguishes genuine reasoning from superficial heuristics or random guessing.
Additionally, we report the consistency ratio and the accuracy based only on the final generated answer (measured by exact match). 

We also evaluate the model’s generalizability on out-of-domain open-ended benchmarks covering different topics. 
For legal reasoning, we use LexEval~\cite{li2025lexeval} open-ended tasks, focusing on legal question answering and summary generation. 
For medical reasoning, we use PubMedQA~\cite{jin2019pubmedqadatasetbiomedicalresearch}, consisting of open-ended questions derived from biomedical research in PubMed. 
We employ Rouge-L as the evaluation metric for both benchmarks.

\paragraph{Implementation Details}

We use both Qwen-2.5-3B-Instruct and Qwen-2.5-7B-Instruct~\cite{qwen2025qwen25technicalreport} as the backbone model, resulting in \textsc{CLARity}$_{\text{3B}}$ and \textsc{CLARity}$_{\text{7B}}$ respectively.
We use Qwen-2.5-7B-Instruct for the initial pass rate and the consistency reward calculation. 
For data augmentation and consistency evaluation, we utilize DeepSeek-V3~\cite{DeepSeekAI2024DeepSeekV3TR} and implement training with the GRPO~\cite{grpo} algorithm using the Verl framework.

In our two-stage refine-then-monitor training pipeline, we randomly sample 500 instances from the training dataset for stage-1, 
with the remaining data used for stage-2. 
When combining data reformulation with the pipeline, 
we apply it only in Stage-2, 
where we start with the original data and begin to reformulate at step 100 with $\alpha=0.5$. See Appendix~\ref{app:implementation:config} for more training details.

\section{Results and Analysis}
\subsection{Main Results}
We report the main results in Table~\ref{tab:main-result}. 
Baselines, including standard RL and PRM RL, show degraded reasoning consistency across both domains, confirming the risks of applying RL to MCQs and the limitations of using small LLMs directly as PRMs. 
In contrast, \textbf{\textsc{CLARity} achieves substantial gains in both consistency and accuracy}.

Moreover, \textbf{\textsc{CLARity} shows strong generalizability}, with consistent gains on diverse unseen open-ended tasks. 
We further validate this by transforming the original validation set into two alternative formats with unchanged content: 
(1) Iterative Judgment (Iter.), where each option is judged independently, and
(2) Shuffled MCQ (Shuf.), where options are recombined using the method in Section~\ref{method:data_aug}. 
Because options in the original MCQs are often correlated, so identifying one correct choice may suffice to solve the whole question. The new formats block such shortcuts to better test reasoning ability. 
As shown in Table~\ref{tab:main-result-generalizability}, \textsc{CLARity} maintains superior performance across both, which we attribute to our data reformulation strategy, as more diverse data usually leads to better generalization~\cite{huang2025ragrladvancingretrievalaugmentedgeneration, zhouDoesLearningMathematical2025}.

To unpack the mechanisms driving these gains, the following analysis delves into how consistency monitoring via small LLMs sharpens logical deduction and prevents reward hacking, alongside the framework's real-world scalability. 
Furthermore, human evaluations confirm that \textbf{\textsc{CLARity} improves holistic reasoning quality, yielding better professionalism and readability.} 
This demonstrates that small general-purpose LLMs can effectively guide expert model training, producing models that even surpass large commercial systems like GPT-4o~\cite{openai2024gpt4ocard}.

\subsection{How Does \textsc{CLARity} Boost Reasoning?}

In this section, we explore the underlying causes of the inconsistency, and identify the key reason for the improvement brought by the consistency-aware learning paradigm in \textsc{CLARity} framework.

\begin{figure}[t]
\centering
\includegraphics[scale=0.255]{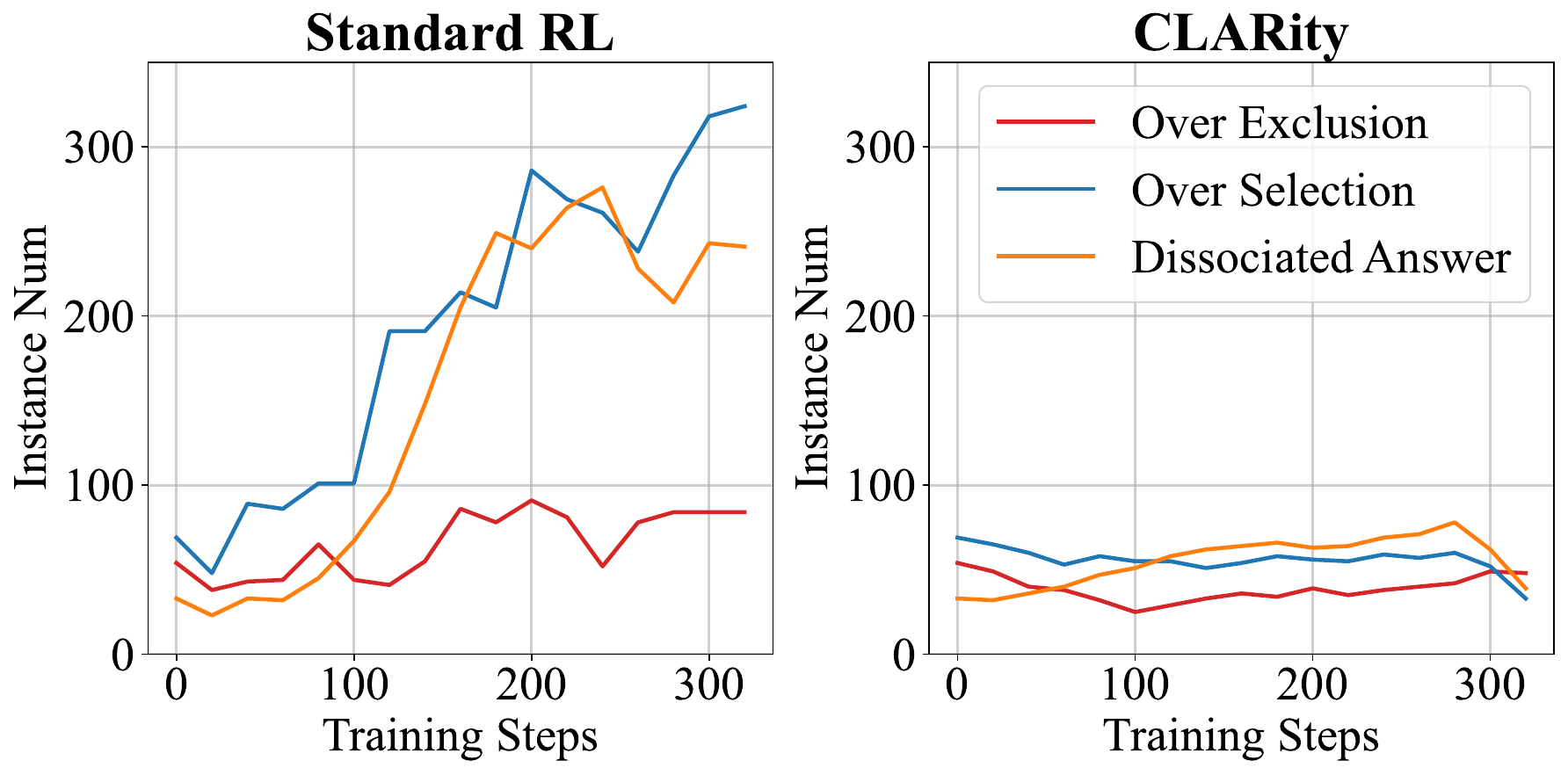}
\caption{Training dynamics of three inconsistency types on Jec-QA validation set.}
\label{fig:discussion:dynamics}
\end{figure}

\begin{table}[t]
\small
\centering
\setlength\tabcolsep{3pt}
\begin{tabular}{l|l|c|c|c}
\toprule
 & \textbf{Type} & \textbf{\# Cons} & \textbf{\# Incons} & \textbf{Cons Rate} \\ 
\midrule
\multirow{2}{*}{\textit{Standard RL}} & Correct & 951 & 285 & 76.9\% \\
 & Incorrect & 513 & 365 & 58.4\% \\
\midrule
\multirow{2}{*}{\textit{\textsc{CLARity}}} & Correct & 1150 & 42 & 96.5\% \\
 & Incorrect & 840 & 82 & 91.1\% \\
\bottomrule
\end{tabular}
\caption{Detailed statistics about consistency and correctness of different methods after RL training. }
\label{tab:analysis:confusion_matrix}
\end{table}

\paragraph{How can inconsistency affect reasoning?} 
To better understand the mechanisms behind inconsistency, we categorize the phenomenon into three types:
\textsc{Over Exclusion} occurs when the reasoning process excludes all available option, yet the final answer selects some options it has just dismissed, suggesting that the model is able to consistently rule out all incorrect options, but it struggles to identify the correct one. 
\textsc{Over Selection} arises when the LLM identifies multiple options as plausible in its reasoning, but the final answer contains only a partial subset of them. 
This points to the model can successfully identify and validate all the correct options but fails to verbalize the exclusion of some wrong options. 
\textsc{Dissociated Answer} represents where the reasoning process explicitly concludes some options are the correct choice, but the final output exist a different, unrelated option. 
This indicates a severe breakdown between the intermediate reasoning and the final answer generation, indicating that the process of excluding irrelevant options and identifying the answers are both unreliable.

\paragraph{Vanilla RL training confuses the model's decision in excluding incorrect options, and causes hesitation on more harder tasks. } 

Building upon our taxonomy, we now analyze how these inconsistencies evolve during training.
We use DeepSeek-V3 as the annotator to track the outputs from the model undergoing standard RL training.
As shown in Figure~\ref{fig:discussion:dynamics}, 
the breakdown by type shows that the overall increase in inconsistencies is primarily driven by a significant rise in \textsc{Over Selection} and \textsc{Dissociated Answer} cases, while the \textsc{Over Exclusion} case increase only marginally, 
suggesting that the primary failure is the model's growing reluctance to exclude wrong cases. 
Moreover, when examining consistency alongside accuracy (Table \ref{tab:analysis:confusion_matrix}), we find that responses to incorrectly-answered questions exhibit much lower consistency than those correct ones. 
This suggests that standard RL makes the model more hesitant when facing difficult questions, 
failing to perform complex reasoning and thus offering very limited improvement in its actual reasoning capability.

\paragraph{\textsc{CLARity} sharpens the model's capacity to exclude incorrect choices.} 
As shown in Figure~\ref{fig:discussion:dynamics} and Table~\ref{tab:analysis:confusion_matrix}, by monitoring response consistency during training, all three types of inconsistency show a significant decrease compared to vanilla RL training, confirming our approach's effectiveness, 
especially in sharpening the model's capacity to exclude incorrect choices compared to the standard RL. 
We attribute these improvements to two key components: the consistency-aware learning mechanism (see discussion below) and the data reformulation pipeline (see Appendix~\ref{app:analysis:ablation:data_reformulation}).

\subsection{Can Small Models Guide Experts?}
Training expert-level LLMs typically requires large models like GPT-4 or additional training to augment domain knowledge~\cite{chen2024huatuogpto1medicalcomplexreasoning, su2025crossingrewardbridgeexpanding}. 
These methods are resource-intensive, requiring considerable financial investment or huge amount of expert-annotated data. 
We investigate whether smaller, general-purpose LLMs can effectively guide the training for expert model.

\begin{table}[t]
\small
\centering
\setlength\tabcolsep{3pt}
\begin{tabular}{l|cc|cc|c}
\toprule
 & \multicolumn{2}{c|}{\textbf{Legal}} & \multicolumn{2}{c|}{\textbf{Medical}} & \textbf{Avg.}\\
 & Acc$^{+}$ & Cons\% & Acc$^{+}$ & Cons\% & Acc$^{+}_{\text{Cons\%}}$ \\ 
\midrule
Qwen2.5 & 28.5 & 92.3 & 42.5 & 88.7 & $35.5_\text{90.5}$ \\
\midrule
\multicolumn{6}{l}{\textit{Open-Ended Rewarding}} \\
\midrule
StrMatch RL & 25.6 & 90.6 & 36.0 & 89.1 & $30.8_\text{89.9}$ \\
PRM RL & 26.1 & 86.5 & 39.8 & 91.2 & $33.0_\text{88.9}$ \\
\midrule 
\multicolumn{6}{l}{\textit{MCQ Rewarding}} \\
\midrule
StrMatch RL & 45.0 & 69.3 & 57.8 & 81.3 & $51.4_\text{75.3}$ \\
PRM RL & 42.0 & 68.2 & 60.2 & \textbf{96.4} & $51.1_\text{82.3}$ \\
\textsc{CLARity} & \textbf{54.4} & \textbf{94.1} & \textbf{63.3} & 89.5 & $\textbf{58.9}_\text{\textbf{91.8}}$ \\
\bottomrule
\end{tabular}
\caption{Different reward modeling methods comparison. \textit{StrMatch} denotes assigning a reward only when the generated answer exactly matches the ground truth.}
\label{tab:analysis:PRM}
\end{table}

\paragraph{Small-scale LLMs fail as PRM.} 
We first examine whether smaller open-source LLMs (Qwen2.5-7B-Instruct) can directly serve as the PRM, monitoring the reasoning process with the same instructions used for larger models. Responses are rewarded only if they are free from any possible error. 
As shown in Table~\ref{tab:analysis:PRM}, we observe two key points. 
First, treating questions as open-ended generation yields significantly lower accuracy, demonstrating that the non-deterministic nature of open-ended answers leads ineffective training. MCQs, with clear and deterministic answer, can provide more stable and generalizable improvements. 
Second, the 7B model fail to serve as an effective MCQ PRM, leading to results substantially inferior to \textsc{CLARity}. 
This is likely due to a combination of limited domain expertise and inadequate model capacity, which together constrain its ability to accurately assess reasoning quality.

\begin{figure}[t]
\centering
\includegraphics[scale=0.6]{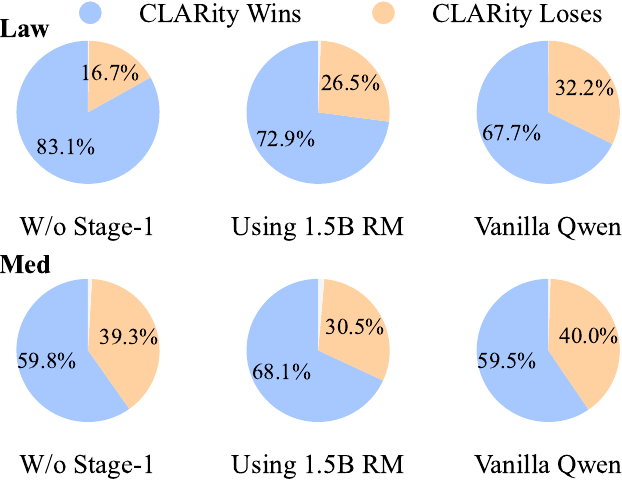}
\caption{Reasoning quality comparison between \textsc{CLARity} and different baselines: without Stage-1, using Qwen-1.5B as the consistency reward model, and the vanilla Qwen2.5-7B-Instruct.}
\label{fig:analysis:cot_quality}
\end{figure}

\paragraph{\textsc{CLARity} enables guidance via monitoring consistency.}
\label{analysis:reward_desigh}
It introduces a consistency-aware learning paradigm, allowing smaller LLMs to monitor response consistency in MCQs. 
As shown in Table~\ref{tab:analysis:PRM}, 
using only a 7B-scale LLM as its consistency checker, \textsc{CLARity} significantly outperforms standard RL and PRM methods, 
and successfully cultivates an expert model with superior reasoning quality and accuracy. 
Further human evaluations in Section~\ref{analysis:human_eval} confirm that \textsc{CLARity} exhibit greater professionalism in their reasoning and are easier for readers to understand.
These results demonstrate that small LLMs, when used to monitor consistency, can effectively curate expert LLMs.

We further investigate whether tiny-sized LLMs ($\le$1.5B, deployable on consumer-grade GPUs) can also monitor reasoning consistency. 
First, we benchmark their inconsistency detection ability against DeepSeek-V3 as a proxy for ground truth\footnote{We validate the reliability of using DeepSeek-V3 as a proxy for ground truth in Appendix~\ref{app:human_eval}.}. 
The detection rates are substantially lower for tiny LLMs: 21\% for Qwen2.5-0.5B and 40\% for Qwen2.5-1.5B, compared to 94\% for the 7B model. 
This indicates that the foundational semantic capabilities of these tiny models are insufficient even for the simpler consistency-checking task. 
Then we experiment with Qwen2.5-1.5B as the consistency reward model within the \textsc{CLARity} framework, 
as shown in Figure~\ref{fig:analysis:cot_quality}, using tiny-sized reward models results in poorer response quality, 
as they encourage simpler, more superficial outputs that the weak consistency checker can easily parse, potentially leading to reward hacking.

\subsection{How to Mitigate Reward Hacking?}
\label{analysis:two_stage}

Reward hacking may occur when tiny-sized LLMs are used as reward models, as their limited capacity prevents them from detecting inconsistencies. 
LLMs with relatively larger size, such as the 7B models, can identify inconsistencies, but they may still lack domain-specific knowledge, preventing them from distinguishing between substantive reasoning and superficially plausible shortcuts, which can also result in reward hacking. 
We propose a 2-stage refine-then-monitor pipeline in \textsc{CLARity} to mitigate this, which we evaluate through experiments to demonstrate its necessity.

\begin{figure}[t]
\centering
\includegraphics[scale=0.385]{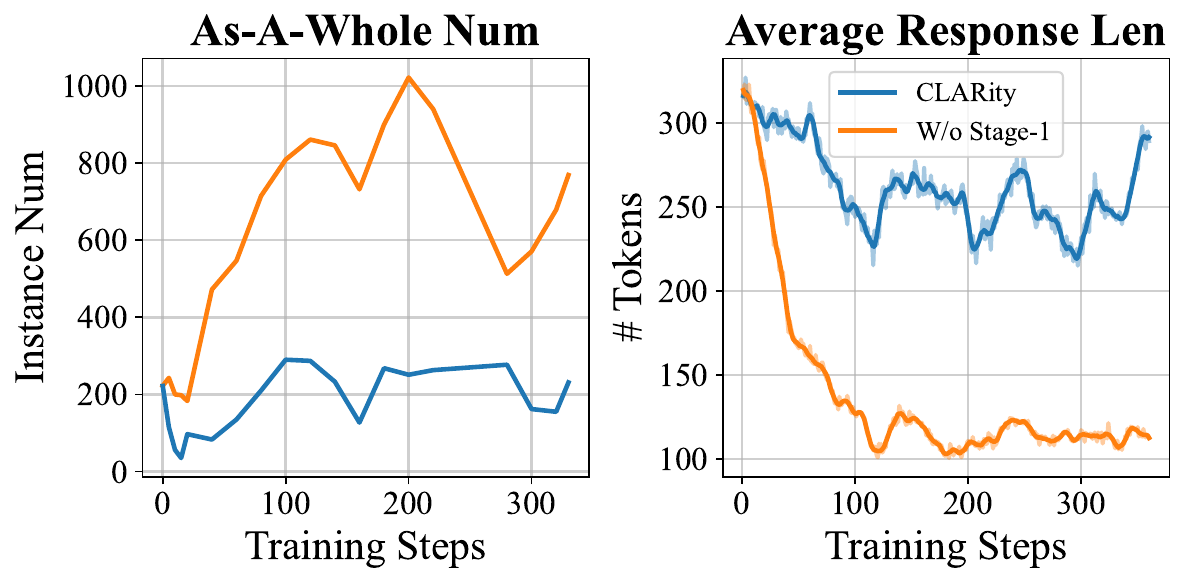}
\caption{Training dynamics of \textit{as-a-whole} response frequency and response length. In \textsc{CLARity}, the first 15 steps correspond to Stage-1.}
\label{fig:discussion:reward_hacking}
\end{figure}

\paragraph{Refine-then-monitor pipeline offers flexible reasoning and avoids hacking.} 
Our 2-stage refine-then-monitor pipeline first encourages explicit reasoning for each option and then uses a consistency reward model to monitor the responses. 
We categorize the responses into two types based on the observed patterns in model reasoning. 
\textit{Explicit}: The reasoning includes an explicit and detailed analysis of options, typically appears in complex questions requiring diverse knowledge. 
\textit{As-a-Whole}: The reasoning is generalized and presented without itemizing options, often uses in simpler questions requiring minimal thought. 

As shown in Figure~\ref{fig:discussion:reward_hacking}, in stage-1, the number of \textit{As-a-Whole} responses decreases as we incentivize a transparent and structured format, 
making the reasoning process legible and easy for the consistency checker to evaluate accurately.
In stage-2, we remove this structural constraint and monitor response quality with the consistency reward. 
We observe a gradual increase in \textit{As-a-Whole} responses, which finally stabilizes. 
This demonstrates that the model has not merely memorized a format but has learned to flexibly apply the appropriate reasoning pattern based on task complexity, achieving consistently high-quality outputs.

\paragraph{Ablating the pipeline leads to reward hacking.}
To prove the necessity of our design, we ablated the pipeline by removing the first stage, 
applying the reward LLM throughout the training. 
We assess response quality using LLM-as-a-Judge~\cite{gu2025surveyllmasajudge}, evaluating answer rationality, professionalism, coherence, clarity, and ease of understanding. 

As shown in Figures~\ref{fig:analysis:cot_quality} and~\ref{fig:discussion:reward_hacking}, removing the first stage leads to a significant degradation in response quality with much shorter responses. 
This is because the trained model quickly learns that simpler, more superficial reasoning is preferred. 
And it may even attempt to conceal complex reasoning to avoid penalties, resulting in reward hacking. 
This hypothesis is further supported by the statistic in Figure~\ref{fig:discussion:reward_hacking}, where the number of \textit{As-a-Whole} responses significantly increases in training, indicating that the model increasingly favor over-simplified reasoning, even when the questions demand substantial analysis. 
See Appendix~\ref{app:casestudy} for case study and Appendix~\ref{app:analysis:reward_model} for further ablation on reward models.

\subsection{\textsc{CLARity} Scales Efficiently and Robustly in Practice}

Recent reasoning models excel in STEM but struggle in specialized fields. As shown in Table \ref{tab:app:robustness_comparison}, DeepSeek-R1-Distill-Qwen underperforms the vanilla Qwen model, indicating that STEM-focused reasoning capability transfers poorly to non-STEM domains.
\textsc{CLARity} addresses this by enabling domain-specific optimization with only a few thousand instances, making it highly practical for data-scarce applications. 
Furthermore, our framework remains highly robust across settings. 
Varying the reformulation rate $\alpha$ or data subsets yields stable performance. 
Notably, reformulating the easiest instances (the vanilla setting) outperforms random or hardest-instance baselines, confirming the value of dynamically optimizing data utility.

Beyond robustness, \textsc{CLARity} minimizes the computational overhead of Process Reward Models, which usually require expensive APIs or extensive annotated data. Instead, it uses a 7B general-purpose LLM for consistency rewards. With asynchronous inference techniques (e.g., \textit{asyncio}), a training step takes about 3.5 minutes—compared to 2 minutes for standard RL—establishing \textsc{CLARity} as a highly scalable, cost-effective solution for resource-constrained scenarios.

\begin{table}[t]
  \small
  \centering
  \renewcommand{\arraystretch}{1.05}
  \setlength\tabcolsep{6.9pt}
  \begin{tabular}{lc|lc}
  \toprule
    \multicolumn{2}{l|}{\textit{General Models}} & \multicolumn{2}{l}{\textit{Reasoning Models}}  \\
    \midrule
    Qwen2.5-7B & 29.3 & DS-Qwen-7B & 16.6  \\
    GPT-4o & 39.9 & o1-preview & 52.3  \\
    \midrule
    \multicolumn{4}{l}{\textsc{CLARity}}  \\
    \midrule
    $\alpha = 0.5$ \textit{(hardest half)} & 53.6 & $\alpha = 0.25$ & 56.2 \\
    $\alpha = 0.5$ \textit{(random half)} & 56.0 & $\alpha = 0.75$ & 53.0 \\
    $\alpha = 0.5$ \textit{(easiest half)}& \textbf{56.4} & $\alpha = 1$ & 51.9 \\
    \bottomrule
  \end{tabular}
  \caption{Performance of \textsc{CLARity} compared to baseline LLMs and across different data reformulation settings on the legal reasoning task.}
  \label{tab:app:robustness_comparison}
\end{table}

\begin{table}[t]
  \centering
  \renewcommand{\arraystretch}{0.9}
  \scalebox{0.8}{
  \begin{tabular}{l|ccc|ccc}
    \toprule
     & \multicolumn{3}{c|}{\textit{Standard RL}} & \multicolumn{3}{c}{\textit{\textsc{CLARity}}} \\
    \midrule
     & Corr. & Prof. & Read. & Corr. & Prof. & Read. \\
    \midrule
    Legal &  3.0 & 2.6 & 2.9 & 3.1 & 2.8 & 2.9 \\ 
    Medical & 3.5 & 3.9 & 3.3 & 3.8 & 4.2 & 3.5 \\
    \midrule
    Average & 3.3 & 3.3 & 3.1 & \textbf{3.5} & \textbf{3.5} & \textbf{3.2} \\
    \bottomrule
  \end{tabular}
  }
  \caption{
    Human evaluation results on both domains.}
  \label{tab:app:human_eval}
\end{table}

\subsection{Human Evaluation}
\label{analysis:human_eval}
We further demonstrate that \textsc{CLARity} can improve the overall response quality beyond reasoning consistency through human evaluation.

\paragraph{Evaluation Metrics}
For both the legal and medical domains, we recruit domain experts who have passed their respective professional qualification exams to assess the quality of 50 randomly sampled responses from the validation set. 
We establish three complementary evaluation criteria, which collectively reflect the overall response quality: 
\textbf{Correctness} (Corr., measuring the logical validity of both the reasoning process and the final conclusion), 
\textbf{Professionalism} (Prof., measuring the factual accuracy and appropriate application of domain knowledge), and 
\textbf{Readability} (Read., measuring the clarity, fluency, and comprehensibility of the generated text). 
We use a 5-point Likert scale for each criterion, where 1 represents "very poor" and 5 represents "very good".

\paragraph{\textsc{CLARity} improves professionalism and readability.}

The results in Table~\ref{tab:app:human_eval} show that \textsc{CLARity} consistently achieves higher average scores than the vanilla RL model, demonstrating its effectiveness in enhancing overall reasoning quality by monitoring reasoning consistency.  See Appendix~\ref{app:human_eval} for more human evaluations. 

\label{app:human_eval:cot_quality}

\section{Related Works}
\paragraph{Reinforcement Learning for LLM}

Reasoning models, such as OpenAI o1~\cite{openai2024openaio1card} and DeepSeek-R1~\cite{guo2025deepseek}, mainly focus on leveraging reinforencement learning methods like PPO~\cite{ppo} and GRPO~\cite{grpo} 
to enhance LLM reasoning abilities.
By rewarding accuracy and format, these methods achieve significant improvement across benchmarks in mathematics~\cite{OpenReasonerZero2025,tinyzero,fengReToolReinforcementLearning2025,yang2025reasonfluxhierarchicalllmreasoning}, logic~\cite{xieLogicRLUnleashingLLM2025}, code generation~\cite{openr1} and multi-hop QA~\cite{song2025r1searcherincentivizingsearchcapability, jin2025searchr1trainingllmsreason, huang2025ragrladvancingretrievalaugmentedgeneration}. 
However, few studies have adapt RL to other domains with only MCQ data available such as law and medicine ~\cite{chen2024huatuogpto1medicalcomplexreasoning, yu2025finemedlmo1enhancingmedicalknowledge}, 
where the clarity of the responses can't be guaranteed. 
We are the first to analyze the effectiveness of RL training in these domains from the perspective of response quality, focusing on logical consistency.

\paragraph{LLM Reasoning Consistency}

refers to whether the model's observed reasoning accurately reflects its internal thinking. 
Many studies highlights LLM consistency as a bottleneck in various settings, including incoherent logical reasoning~\cite{kirchner2024proververifiergamesimprovelegibility, ferreira2025truthfulfabricatedusingcausal,arcuschin2025chainofthoughtreasoningwildfaithful}, resistance to verbalizing hints~\cite{chen2025reasoningmodelsdontsay} or deliberately concealing its true capabilities~\cite{meinke2025frontiermodelscapableincontext}. 
To improve consistency, current research typically employs two ways: 
constructing high-quality supervised fine-tuning (SFT) data~\cite{chen2024consistentnaturallanguageexplanationsexplanationconsistency, paul2024makingreasoningmattermeasuring, ferreira2025truthfulfabricatedusingcausal, chua2025biasaugmentedconsistencytrainingreduces}, 
and filtering low-quality outputs during training~\cite{wang2025chainofprobeexaminingnecessityaccuracy, kirchner2024proververifiergamesimprovelegibility, baker2025monitoringreasoningmodelsmisbehavior}. 
Our work, focusing on improving data quality and dynamically rewarding high-quality responses, extends these approaches with the context of RL training.

\section{Conclusion}
We identify MCQ RL often produces superficially accurate yet inconsistent reasoning. 
We thus introduce \textsc{CLARity}, an efficient RL framework to guide expert models through a consistency-aware refine-then-monitor training pipeline. 
Our method improves data utilization and delivers holistic enhancements in reasoning quality, including accuracy, professionalism, and readability, all without relying on large commercial models or expert-annotated corpora, thus
offering a cost-effective and generalizable solution.

\section*{Limitations}

\paragraph{Limited Exploration of Data Augmentation Approach} Although this work demonstrates the effectiveness of our proposed data reformulation strategy, its exploration remains relatively limited. 
In particular, in this paper, we do not fully exploit its potential flexibility—for example, dynamically varying the number of candidate options per question, or mixing data from multiple distinct domains to create more diverse and challenging training samples. 
We believe these directions hold promise for further improving generalization and encourage future work to investigate them using our proposed data reformulation strategy further.

\paragraph{Limited Integration with More Training Paradigms} While designed as a plug-and-play mechanism, we have not explored integrating \textsc{CLARity} with other popular learning paradigms, such as curriculum learning. Conventional curriculum learning requires large datasets with clearly defined difficulty levels—a stark contrast to the severe data scarcity in our target domains. If more instances with clear difficulty levels were available, we believe our method could benefit a further curriculum learning paradigm to improve training efficiency (e.g., grouping data based on any required difficulty level).

\paragraph{Limited Exploration of Domain Generality} While effective in specialized fields, our evaluation is restricted to two domains (law and medicine) with small validation sets. We have not yet conducted cross-domain tests to verify if \textsc{CLARity} enhances general reasoning in broader, higher-diversity areas. We leave the empirical validation of its cross-domain transferability and broader impact to future work.

\section*{Acknowledgements}
This work is supported in part by Beijing Natural Science Foundation (L253001), Natural Science Foundation of China (92570207) and Noncommunicable Chronic Diseases-National Science and Technology Major Project (2023ZD0512400). We thank the anonymous reviewers for their valuable feedback. 
For any correspondence, please contact Yansong Feng.

\bibliography{custom}

\clearpage

\appendix
\label{sec:appendix}

\section{Detailed Algorithm}

The detailed algorithm of \textsc{CLARity} is shown in Algorithm~\ref{alg:data}. 
$N$ denote the total number of instances in the training dataset $D$, and $M$ the number of candidate options for each multiple-choice question. 
Each problem instance $I_i = (Q_i, C_i, Ans_i) \in D$ consists of a query statement $Q_i$ and $M$ candidate options $C_i := \{C_{i1}, C_{i2}, \dots, C_{iM}\}$, with an answer $A_i \subseteq C_i$, 
$A_i$ can be any subset of $C_i$, including the full set and the empty set.

\section{Human Evaluations}
\label{app:human_eval}

We further validate the effectiveness of our proposed methods through human evaluation. 
The annotator's payment is adequate given the participants’ demographic. 
The annotators are informed of how the data would be used. 

\subsection{Model-Human Agreement}

First we examine the effectiveness of our consistency reward model. 
We recruit a graduate proficient in English and Chinese, and provide him with a sample of 100 instances from the validation set for each domain, use the model trained with vanilla RL to generate response, and ask him to evaluate the consistency between reasoning trajectory and answer. 
The Pearson correlation coefficient between human and DeepSeek-V3 is 0.610 with $p=4.2 \times 10^{-10}$ averaged on two domains, demonstrating LLMs can effectively detect inconsistencies, with a relatively high correlation between model and human.

\subsection{Quality Evaluation}

\paragraph{Training Data Quality Evaluation}
We conduct human evaluation on training data to further validate its quality. 
We randomly sample 50 samples from the augmented dataset obtained from Algorithm~\ref{alg:data} for each domain, 
and examine whether the data quality from two aspects: 
whether the statement natural, fluent, and easy for humans to understand, 
and whether the key information in the content between the propositions and original data is the same (i.e. The augmented problem is a well-defined reformulation of the original problem.) 
We find all 50 samples are well-presented without any incoherence, 
and 49 in 50 samples maintains all the key information compared with the original data.
See Appendix~\ref{app:implementation:dataset} for training dataset examples. 
All training data are used in accordance with their original intent and licenses.

\paragraph{Response Quality Evaluation}

We provide the details of evaluation metrics below. See evaluation results and discussion in Section~\ref{analysis:human_eval}.

Correctness (5 points):
Focuses on whether the reasoning and conclusions are logically sound and internally consistent. The emphasis is on logical validity rather than domain-specific factual precision.

Professionalism (5 points):
Evaluates the factual accuracy and appropriate knowledge application. The key concern is whether the cited knowledge is correct and meaningfully contributes to the reasoning process.

Readability (5 points):
Assesses how clearly and smoothly the response can be read and understood, reflecting overall coherence and linguistic fluency.


\section{More Implementation Details}

\subsection{Training Configuration}
\label{app:implementation:config}
When using our proposed 2-stage training pipeline with consistency reward, 
for all experiments we uniformly samples 500 instances from the training dataset for stage 1, training for 2 epochs with only answer reward and format reward, and use the rest instances for stage-2 training, with 8 epochs using our consistency reward, answer reward and format reward.
Additionally, we change the training data into our shuffled and diversified dataset in stage-2 training after 100 steps when using our data augmentation approach.

The training was conducted on a single node equipped with 8 A800 GPUs using the veRL library~\cite{Sheng_2025}. The key training configuration is shown below in Table~\ref{tab:app:implementation:config}. 
The detailed reward weight is listed in Table~\ref{tab:app:implementation:weight}. Note that we do not conduct hyperparameter searching, so the current performance is underoptimal, and has the potential of further imperovement.

\begin{table}[ht]
\centering
\small
\setlength\tabcolsep{2.68pt}
\begin{tabular}{llc}
\toprule
\textbf{Category} & \textbf{Parameter} & \textbf{Value} \\
\midrule
\multirow{6}{*}{General} 
  & Advantage estimator           & GRPO  \\
  & Gamma ($\gamma$)                & 1  \\
  & Lambda ($\lambda$)              & 1  \\
  & Batch size                   & 128  \\
  & Max prompt length            & 1024  \\
  & Gradient checkpointing       & Enabled  \\

\midrule
\multirow{10}{*}{Actor} 
  & Learning rate                   & $1 \times 10^{-6}$  \\
  & Mini-batch size                 & 1024  \\
  & Dynamic batch size              & Enabled  \\
  & KL penalty role                    & Loss \\
  & KL loss type                     & Low-variance KL \\
  & KL loss coefficient ($\beta$)    & 0.001  \\
  & Entropy coefficient             & 0.001  \\
  & Clip ratio                       & 0.2  \\
  & Gradient clipping                & 1.0  \\
  & Sequence parallel size           & Model-specific  \\

\midrule
\multirow{6}{*}{Rollout} 
  & Backend                          & vLLM  \\
  & Tensor model parallel size       & 2  \\
  & Rollouts per sample              & 6  \\
  & Nucleus sampling $p$                          & 1.0  \\
  & GPU memory utilization           & 0.4  \\
  & Sampling temperature            & 1.0  \\

\midrule
\multirow{1}{*}{Qwen2.5-7B} 
  & Max response length             & 2048  \\
\bottomrule
\end{tabular}
\caption{Training configurations. We use VeRL 0.4.1.dev version.}
\label{tab:app:implementation:config}
\end{table}

\begin{table}[ht]
\centering
\small
\setlength\tabcolsep{2.68pt}
\begin{tabular}{llcc}
\toprule
\textbf{Phase} & \textbf{Value$_{True}$} & \textbf{Value$_{False}$} & \textbf{Weight}\\
\midrule
\multicolumn{3}{l}{\textbf{\textit{Answer Reward}}} \\
\midrule
Vanilla RL & 1.0 & 0.0 & 1.0 \\ 
stage 2 & 1.0 & 0.0 & 1.0 \\
\midrule
\multicolumn{3}{l}{\textbf{\textit{Format Reward}}} \\
\midrule
Vanilla RL & 1.0 & 0.0 & 1.0 \\ 
stage 1 & 1.0 & 0.0 & 1.0 \\
stage 2 & 0.0 & -1.0 & 0.1 \\
\midrule
\multicolumn{3}{l}{\textbf{\textit{Structure Reward}}} \\
\midrule
Stage-1 & 1.0 & 0.0 & 1.0 \\
\midrule
\multicolumn{3}{l}{\textbf{\textit{Consistency Reward}}} \\
\midrule
Stage-2 & 0.0 & -1.0 & 1.0 \\
\bottomrule
\end{tabular}
\caption{Reward weights. The final reward score equals to Weight*Value$_{True/False}$}
\label{tab:app:implementation:weight}
\end{table}

\subsection{Prompt and Dataset Example}
\label{app:implementation:dataset}
\label{app:implementation:prompt}

\paragraph{Data Augmentation}
We provide the prompts used in augmenting the original dataset, and the instance example of the original dataset and the augmented dataset. 
The original dataset example of Jec-QA is shown in Table~\ref{tab:app:implementation:dataset:jecqa}, and MedQA in Table~\ref{tab:app:implementation:dataset:medqa}. 
Then we concatenate the original statement with options and conduct polish and diversify using LLM. 
Table~\ref{tab:app:implementation:prompt:polish} shows the prompt used for data reformulation.
The final data example after our pipeline is shown in Table~\ref{tab:app:implementation:dataset:jecqa_proposition} and Table~\ref{tab:app:implementation:dataset:medqa_proposition}.

\paragraph{2-stage Consistency Training and Evaluation}
During 2-stage consistency training, we use a special reward model to monitor the CoT consistency. 
We use 3-shot in-context learning to instruct the LLM to parse the believed-correct options in the CoT using the Vllm library~\cite{kwon2023efficient}. 
The prompt for the consistency reward model is shown in Table~\ref{tab:app:implementation:prompt:consistency_reward}. 
We also provide the LLM-as-a-judge prompt used in evaluating the quality of the generated CoTs in Table~\ref{tab:app:implementation:prompt:judgement}.

\section{Additional Experimental Results}

\subsection{Ablation Study on Data Reformulation}
\label{app:analysis:ablation:data_reformulation}
As shown in Table~\ref{tab:main-result} and Table~\ref{tab:main-result-generalizability}, 
the proposed dynamic data reformulation approach significantly improves model accuracy across various formats and content. We conduct ablations to assess the impact of: (1) LLM-based diversified data refinement (training with only string-level concatenated data), and (2) pass-rate-based reformulation (using original data without reformulation, or using randomly deconstructed data). The results, shown in Table~\ref{app:analysis:ablation:data_reformulation}, reveal that removing LLM-based refinements decreases performance, underscoring the importance of using LLMs to eliminate redundant expressions and diversify content. Additionally, reformulating easy data during training yields the best performance, confirming the effectiveness of our method and supporting prior work on introducing more challenging instances in RL training~\cite{Polaris2025, huang2025ragrladvancingretrievalaugmentedgeneration, li2025limrrlscaling}.

Notably, consistency also improves with reformulation compared to standard RL, suggesting that curating more challenging data mitigates inconsistency, as tasks requiring complex reasoning naturally encourage models to reason more effectively.

\begin{table}[t]
\small
\centering
\setlength\tabcolsep{3pt}
\begin{tabular}{ll|c|c|c|c}
\toprule
 & & \%\textbf{Pass Rate}  & \textbf{Acc$^+$} & \textbf{Cons} & \textbf{Acc} \\ 
\midrule
Standard RL & &  \_ & 51.4 & 75.3 & 61.1 \\
 \midrule
\textit{+ Random} & Data$_\text{LLM}$ & 22.0 & 49.7 &  80.0 &  60.1    \\
\midrule
\multirow{2}{*}{\textit{+ Pass-Rate}} & Data$_\text{StrCat}$ & 13.8 & 53.6 & 78.8 & 59.0 \\
 & Data$_\text{LLM}$ & 14.5 & \textbf{54.9} & \textbf{82.2} & \textbf{61.1} \\
\bottomrule
\end{tabular}
\caption{Ablation study on data reformulation pipeline. \textit{\%Pass Rate} denotes the initial pass rate on Qwen2.5-7B-Instruct.
}
\label{tab:app:analysis:training_pipeline}
\end{table}

\subsection{Detailed Training Dynamics}
\begin{figure*}[t]
\centering
\includegraphics[width=2.1\columnwidth]{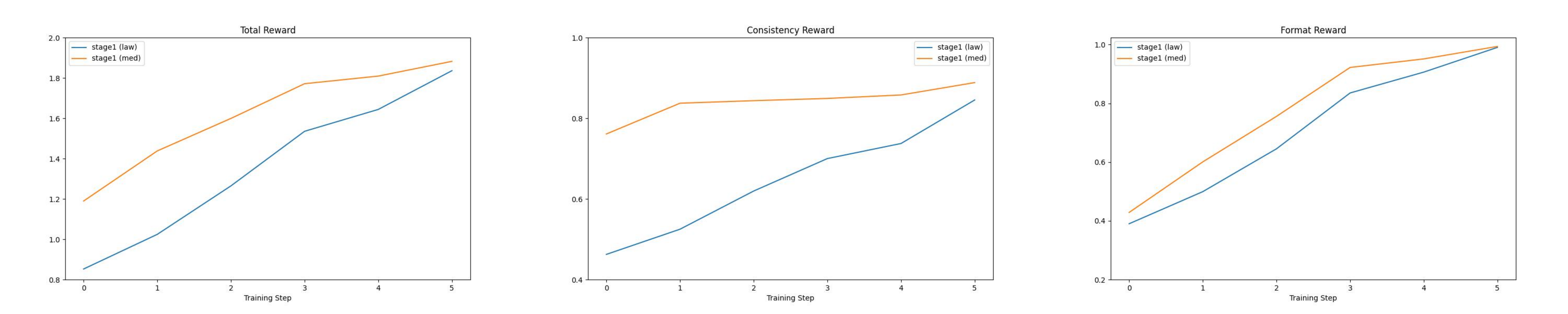}
\caption{Training dynamics of \textsc{CLARity} in Stage-1.}
\label{fig:app:results:dynamics_stage1}
\end{figure*}

\begin{figure*}[t]
\centering
\includegraphics[width=1.5\columnwidth]{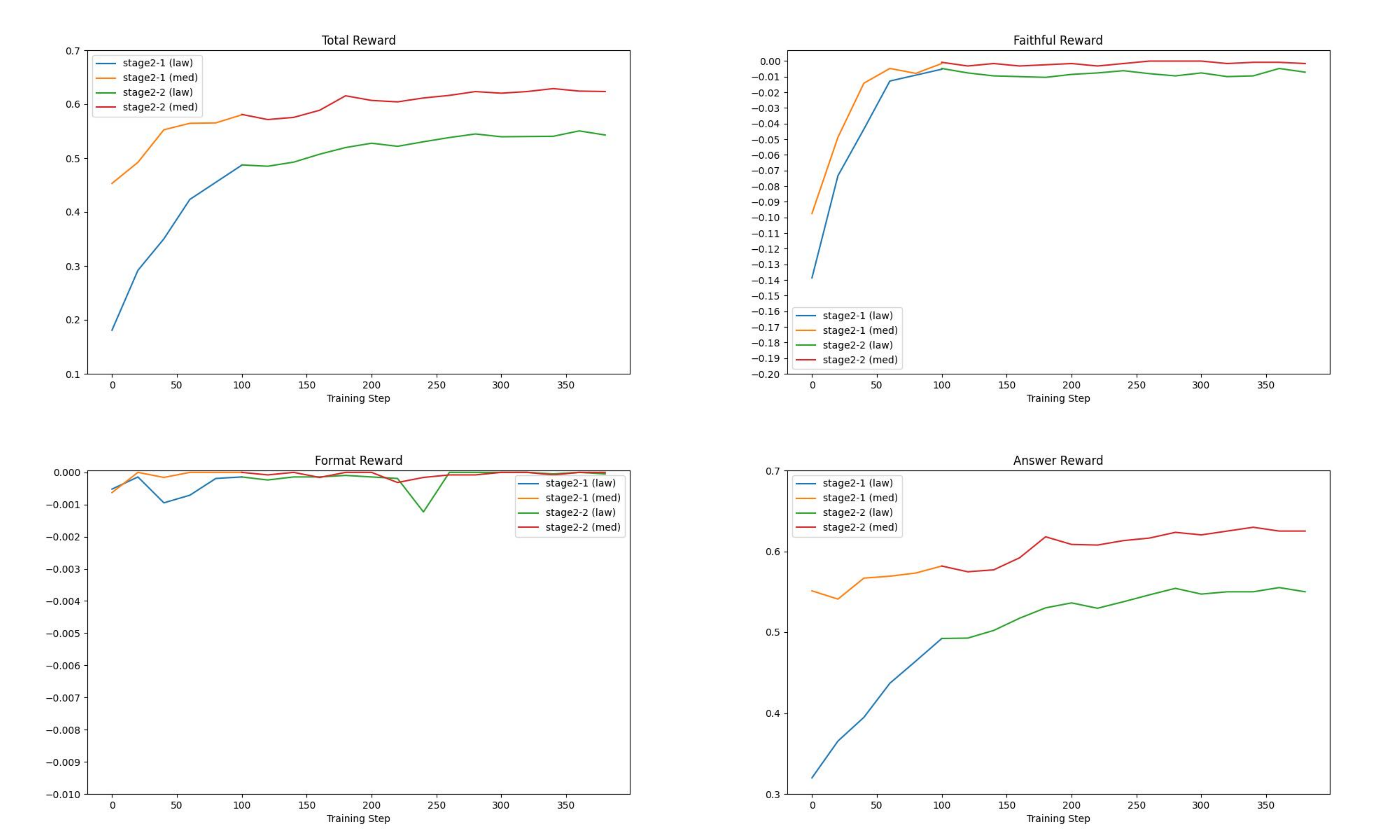}
\caption{Training dynamics of \textsc{CLARity} in Stage-2.}
\label{fig:app:results:dynamics_stage2}
\end{figure*}

We present the more detailed dynamics in RL training. 
In Figures~\ref{fig:app:results:dynamics_stage1} and \ref{fig:app:results:dynamics_stage2}, we report the log of rewards during training, including the answer reward, the consistency reward and the format reward.

\subsection{Results For Rule-Based Consistency Reward Model} 
\label{app:analysis:reward_model}

We explore the feasibility of using a rule-based consistency reward model that leverages exact pattern matching of regular expressions.

We find using the rule-based consistency reward results in a performance drop of 11.4\% in answer accuracy on the validation set,
and shows almost no improvement in out-of-domain tasks. 
The rule-based approach forces the model to generate responses with a fixed format designed to match predefined patterns rather than encouraging deeper reasoning, 
which makes the model more prone to overfitting during training, producing shorter responses that overly focused on matching the regular expression.

\section{Case Study}
\label{app:casestudy}
We present case studies from both domains in Table~\ref{tab:app:casestudy:jecqa} and Table~\ref{tab:app:casestudy:medqa}.
The vanilla Qwen2.5-7B-Instruct model already demonstrates basic legal and medical reasoning capabilities, allowing it to perform limited analytical reasoning. However, its accuracy remains low, often relying on incorrect or incomplete knowledge. Moreover, its responses contain redundant and shallow analyses, suggesting that the distilled reasoning patterns it learned are largely superficial—capturing only surface structures rather than genuine problem-understanding or reasoning depth.

After reinforcement learning, the model’s accuracy improves. However, as illustrated in the tables, the standard outcome-based RL introduces inconsistencies: the model may reject a candidate option during reasoning but later select it as the final answer. This inconsistency highlights that outcome-based rewards alone are insufficient to ensure high-quality reasoning.

\textsc{CLARity} address this issue through a consistency-aware mechanism that explicitly monitors reasoning alignment between intermediate judgments and final conclusions. Additionally, its two-stage refine-then-monitor pipeline further enhances response quality, mitigating potential reward hacking that arises when small-scale reward models fail to detect inconsistencies.
When the refine-then-monitor stage is removed (\textit{W/o Stage-1} in the tables), reasoning quality significantly declines, as the model tends to favor oversimplified reasoning patterns even for questions requiring substantial analytical depth.

\begin{table*}
\small
\begin{tcolorbox}[
    colback=gray!5,
    colframe=gray!60!black,
    title=Data Polishing Prompt Template,
    fonttitle=\bfseries,
    sharp corners,
    boxrule=1pt,
]
I am a \{legal/medical\} school professor preparing final exam questions for my \{legal/medical\} course. I would like to design a statement analysis question, where students are required to judge whether the given statement is correct and provide supporting reasoning. To do this, I select a question stem from my multiple-choice question bank and combine it with one of its answer options to transform it into a statement for analysis. However, the stem may still contain phrasing specific to multiple-choice format (e.g., “Which of the following is …?”) or redundant punctuation, and may result in illogical reasoning. Therefore, I need your help to polish the statement: remove any phrasing related to multiple-choice format, adjust sentence structure or add connecting words if necessary, and ensure that the final version is logically coherent and easy to understand. You should also diversify the statement by adjusting word order, replacing synonyms or conjunctions, or introducing fictional names and places that do not alter the meaning. This will help diversify the question bank. Please enclose the polished statement within curly braces.

\vspace{0.5em}
\texttt{\{Examples\}}

\vspace{1em}
\hrule
\vspace{1em}

\textbf{Original statement:} \texttt{\{Original\_statement\}}
\vspace{5pt}
\end{tcolorbox}
\caption{Prompt for dynamic data reformulation. The original prompt for Jec-QA is written in Chinese, because Jec-QA is presented in Chinese.}
\label{tab:app:implementation:prompt:polish}
\end{table*}

\begin{table*}
\small
\begin{tcolorbox}[
    colback=gray!5,
    colframe=gray!60!black,
    title=Consistency Parsing Prompt Template,
    fonttitle=\bfseries,
    sharp corners,
    boxrule=1pt,
]
You are a \{legal/medical\} expert. You will be given a multiple‑choice \{legal/medical\}‑exam question and a student's line of reasoning. Based on the student's reasoning, identify the final answer selected by the student. Follow the format in the examples below and enclose the answer the student believes is correct in curly braces, e.g. "{{}}", "{{B}}" or "{{ABD}}".

\vspace{0.5em}
\texttt{\{Examples\}}

\vspace{1em}
\hrule 
\vspace{1em}

Now, based on the examples above, analyze the following question and the student's reasoning, and state the answer the student believes is correct: \\
\textbf{Question:} \texttt{\{Question\}} \\
\textbf{Reasoning:} \texttt{\{Chain-of-Thought\}}

\vspace{5pt}
\end{tcolorbox}
\caption{Prompt for parsing the answer for consistency reward model. The original prompt for Jec-QA is written in Chinese, because Jec-QA is presented in Chinese.}
\label{tab:app:implementation:prompt:consistency_reward}
\end{table*}

\begin{table*}
\small
\begin{tcolorbox}[
    colback=gray!5,
    colframe=gray!60!black,
    title=LLM-as-a-judge Prompt Template,
    fonttitle=\bfseries,
    sharp corners,
    boxrule=1pt,
]
You are a medical expert. Below, you will find a medical exam question along with two responses provided by two candidates. Please compare the quality of these two responses based on various factors, including but not limited to: the rationality of the answer, readability, logical coherence, clarity, and ease of understanding.

Please evaluate both answers and determine which one is of higher quality. Use your expertise to carefully analyze the responses. Your final answer should highlight the better response by wrapping it in curly braces, either \"{{Answer 1}}\" \ or \"{{Answer 2}}.\" \\

\vspace{1em}
\hrule 
\vspace{1em}

\textbf{Question:} \texttt{\{Question\}} \\
\textbf{Answer 1:} \texttt{\{answer\_1\}} \\
\textbf{Answer 2:} \texttt{\{answer\_2\}} \\

\vspace{5pt}
\end{tcolorbox}
\caption{Prompt for LLM-as-a-judge.}
\label{tab:app:implementation:prompt:judgement}
\end{table*}

\begin{CJK}{UTF8}{gbsn}
\begin{table*}[ht]
\small
\centering
\begin{tabular}{c}
\toprule
\multicolumn{1}{p{1.96\columnwidth}}{问题:
甲公司与乙公司就双方签订的加工承揽合同达成仲裁协议，约定一旦合同履行发生纠纷，由当地仲裁委员会仲裁。后合同履行发生争议，甲公司将乙公司告上法庭。对此乙公司没有向受诉法院提出异议。开庭审理中，甲公司举出充分证据，乙公司败诉几成定局，于是乙公司向法院提交了双方达成的仲裁协议。法院审查后认为该仲裁协议无效，此时应如何处理?

\texttt{Question:}
Company A and Company B signed a processing contract and reached an arbitration agreement, stipulating that any disputes arising from contract performance would be arbitrated by the local arbitration commission. Later, a dispute arose during contract performance, and Company A sued Company B in court. Company B did not raise any objections to the court's jurisdiction. During the trial, as Company A presented sufficient evidence, making Company B almost certain to lose, Company B submitted the arbitration agreement to the court. After reviewing it, the court found the arbitration agreement invalid. How should the case be handled at this point?
} \\
\midrule
\multicolumn{1}{p{0.94\textwidth}}{\texttt{Options:}
\textbf{A}: 继续审理 \hfill \textbf{A}: Continue the trial.

\textbf{B}: 判决该仲裁协议无效\hfill \textbf{B}: Declare the arbitration agreement invalid.

\textbf{C}: 如甲公司对仲裁协议效力没有异议，则裁定驳回起诉 \hfill \ \textbf{C}: If Company A does not object to the validity of the arbitration agreement, dismiss the lawsuit by ruling.

\textbf{D}: 将仲裁协议的效力问题移交有关仲裁委员会审理  \hfill \  \textbf{D}: Transfer the issue of the arbitration agreement’s validity to the relevant arbitration commission for review.
}
\\
\bottomrule
\end{tabular}
\caption{An example of Jec-QA case analysis questions. Solving such question requires the model to clarify legal relationships, identify applicable laws, and conduct comprehensive reasoning in complex scenarios rather than rote memorization.}
\label{tab:app:implementation:dataset:jecqa}
\end{table*}
\end{CJK}

\begin{table*}
\centering
\small
\setlength{\tabcolsep}{5pt}
\begin{tabular}{p{2.0\columnwidth}r}
\toprule
\texttt{Question:} A 23-year-old man comes to the physician for evaluation of decreased hearing, dizziness, and ringing in his right ear for the past 6 months. Physical examination shows multiple soft, yellow plaques and papules on his arms, chest, and back. There is sensorineural hearing loss and weakness of facial muscles bilaterally. His gait is unsteady. An MRI of the brain shows a 3-cm mass near the right internal auditory meatus and a 2-cm mass at the left cerebellopontine angle. The abnormal cells in these masses are most likely derived from which of the following embryological structures? \\
\midrule
\texttt{Options:} \textbf{A}: Neural tube \ \ \ \ \  \textbf{B}: Surface ectoderm \ \ \ \ \  \textbf{C}: Neural crest \ \ \ \ \  \textbf{D}: Notochord \ \ \ \ \  \textbf{E}: Mesoderm \\
\bottomrule
\end{tabular}
\caption{An example of MedQA-USMLE questions. These questions assess the model's ability to apply knowledge, concepts, and principles, and the ability to demonstrate fundamental patient-centered skills.}
\label{tab:app:implementation:dataset:medqa}
\end{table*}

\begin{CJK}{UTF8}{gbsn}
\begin{table*}[]
\small
\centering
\begin{tabular}{p{0.8\columnwidth}p{1.2\columnwidth}}
\toprule
\multicolumn{2}{p{2\columnwidth}}{\textbf{问题:} 关于民事诉讼法、行政法与行政诉讼法等法律法规的相关内容，下列选项中错误的说法是？

\textbf{Question:} On matters related to the Civil Procedure Law, the Administrative Law, the Administrative Litigation Law, and other relevant laws and regulations, which of the following statements is incorrect? } \\
\midrule
选项：  \textbf{A}: 5月18日，某市第三中学发生学生集体食物中毒事件，根据《突发公共卫生事件应急条例》的相关规定，第三中学在事发后2小时内向市卫生局报告的处理措施符合法律要求。

\textbf{B}: 某商场甲与某电视机生产厂家乙因货款问题产生纠纷，乙遂拒绝向甲商场供货。当客户向该商场销售员询问是否有乙厂的电视机时，销售员故意宣称：“乙厂的产品不合格，返修率高，为保护消费者利益，我商场已拒绝销售乙厂的产品。”根据《反不正当竞争法》相关规定，甲商场销售员的上述行为属于诋毁乙厂商业信誉的不正当竞争行为。

\textbf{C}: 赵某向陈某借一手机使用，后来陈某向赵某讨要时，赵某表示早就已经将手机还给陈某了，两人因此发生了纠纷。陈某诉至镇上法庭，法庭采用简易程序审理了本案。根据相关法律规定，法庭当庭宣判，告知当事人自己来领取裁判文书，法院将不去送达的做法是正确的。

\textbf{D}: 阳光科技公司与星辰制造公司就双方签订的设备加工合同达成仲裁协议，约定若在合同履行过程中出现任何纠纷，均由所在地仲裁委员会进行仲裁。后来在合同执行阶段双方产生争议，阳光科技公司直接将星辰制造公司诉至法院，而星辰制造公司未在法定期间内向受理法院提出管辖权异议。在案件审理过程中，阳光科技公司提供了完整的证据链，星辰制造公司面临败诉风险，此时才向法院提交了双方之前签订的仲裁协议。经法院审查认定该仲裁协议不具备法律效力，在此情况下，法院应当继续审理本案。 &  \texttt{Options:} \textbf{A}. On May 18, a mass food-poisoning incident occurred at a city’s No. 3 Middle School. Under the Regulations on Emergency Response to Public Health Emergencies, the school’s action of reporting to the municipal health bureau within two hours after the incident complies with legal requirements. 

\textbf{B}. A department store (Party A) and a TV manufacturer (Party B) had a payment dispute, and Party B then refused to supply goods to Party A. When a customer asked the store’s salesperson whether the store carried Party B’s televisions, the salesperson deliberately stated: “Party B’s products are substandard and have a high repair rate. To protect consumers’ interests, our store has refused to sell Party B’s products.” Under the Anti-Unfair Competition Law, the salesperson’s conduct constitutes unfair competition by disparaging Party B’s commercial reputation.

\textbf{C}. Zhao borrowed a mobile phone from Chen. Later, when Chen asked for it back, Zhao claimed he had already returned it, leading to a dispute. Chen sued in the town-level people’s court, which tried the case under the summary procedure. The court pronounced the judgment in court and told the parties to come and collect the written judgment themselves, stating that the court would not effect service. According to the relevant laws, this practice is correct.

\textbf{D}. Sunshine Technology Co. and Starlight Manufacturing Co. entered into an arbitration agreement regarding their equipment-processing contract, stipulating that any disputes arising during performance of the contract shall be submitted to the local arbitration commission for arbitration. During performance, a dispute arose and Sunshine Technology directly sued Starlight Manufacturing in court. Starlight did not raise a jurisdictional objection with the court within the statutory period. In the course of the proceedings, Sunshine Technology produced a complete chain of evidence and Starlight faced the risk of losing; only then did Starlight submit the previously concluded arbitration agreement to the court. Upon review, the court determined that the arbitration agreement was not legally valid. In this situation, the court should continue hearing the case. \\
\bottomrule
\end{tabular}
\caption{Data example of Jec-QA after applying our augmentation.}
\label{tab:app:implementation:dataset:jecqa_proposition}
\end{table*}
\end{CJK}

\begin{table*}
\centering
\small
\setlength{\tabcolsep}{5pt}
\begin{tabular}{p{2.0\columnwidth}r}
\toprule
\texttt{Question:} Which of the following statements are wrong: \\
\midrule
\texttt{Options:} \textbf{A}: A startup is working on a novel project in which they claim they can replicate the organelle that is defective in MELAS syndrome. If their project is to mimic the metabolic processes of this organelle, they must be able to replicate fatty acid synthesis.

\textbf{B}: Researchers are experimenting with hormone levels in mice in fasting and fed states. To test hormone levels in the fed state, the mice are given an oral glucose load and various hormones are measured in a blood sample. Researchers are most interested in the hormone whose blood levels track evenly with C-peptide levels. Based on the experimental setup, the hormone the researchers are most interested in is responsible for fatty acid breakdown in the body.

\textbf{C}: James is a 23-year-old man residing in Boston who presents to his physician for evaluation of progressive hearing loss, dizziness, and tinnitus in his right ear over the past six months. During physical examination, multiple soft, yellowish plaques and papules are noted on his arms, chest, and back. Audiometric testing reveals sensorineural hearing loss, and neurological examination demonstrates bilateral facial muscle weakness along with an unsteady gait. Brain MRI reveals two masses: a 3-cm lesion near the right internal auditory meatus and a 2-cm mass at the left cerebellopontine angle. Given these clinical and imaging findings, the abnormal cells comprising these masses are most likely derived from the neural tube.
            
\textbf{D}: Emma is a 45-year-old woman with a known history of hypertension and bipolar disorder. She visits your clinic today due to new-onset tremors, along with complaints of intense thirst and frequent urination. While her bipolar disorder had been well-controlled with her previous medication regimen, she recently introduced a new drug. Given the symptoms she is experiencing, the medication she most likely started is valproate.

\textbf{E}: A 27-year-old man is running on the treadmill at his gym. His blood pressure prior to beginning his workout was 110/72. Based on the physiological response to exercise, a decreased heart rate may be seen in this man now that he is exercising. \\
\bottomrule
\end{tabular}
\caption{Data example of MedQA after applying our augmentation.}
\label{tab:app:implementation:dataset:medqa_proposition}
\end{table*}

\begin{algorithm*}[tb]
\small{
\begin{algorithmic}[1]
   \STATE {\bfseries Input:} Dataset $\mathcal{D}=\{(Q_i, C_i, Ans_i)\}_{i=0}^{N-1}$, where $C_i = \{c_{i1}, \dots, c_{iM}\}$, $Ans_i \subseteq C_i$, shuffle proportion $\alpha \le 1$, statement template set $S$, Reward weight $\alpha_1$, $\beta_1$, $\alpha_2$, $\beta_2$, $\gamma$, shuffle proportion $\alpha_{shuf}$, number of iterations in both stages
   
   \red{// Dynamic Data Reformulation}
   \FOR{each instance $I_i = (Q_i, C_i, Ans_i)$ in $\mathcal{D}$}
        \FOR{each option $c_{ij}$ in $C_i$}
            \STATE $c_{ij}^{Div} = \text{LLM}_{Refine}(\text{StrCat}(Q_i, c_{ij}))$
            \STATE $Correct_{c_{ij}^{Div}} = \mathbb{I}_{\ c_{ij} \in Ans_i}$
        \ENDFOR
    \ENDFOR
    
    \STATE Descending sort dataset $\mathcal{D}$ by Qwen2.5-7B-Instruct initial pass rate  
    \STATE $\mathcal{D}^{\text{Shuffling}} = \mathcal{D}[0 : \alpha N]$
    
    \STATE $C^{\text{Shuffling}} = \bigcup_{i=0}^{\alpha N-1} C_i$

    \red{// Random Grouping}
    \WHILE{$C^{\text{Shuffling}}$ is not empty}
        \STATE $C^{'}$ = Randomly sample $M$ candidates from $C^{\text{Shuffling}}$ without replacement
        \STATE $Ans^{'} = \{ C \mid C \in C^{'}\text{, } Correct(C) = 1   \}$
        \STATE $Q^{'}$ =  Randomly sample a template from $S$
        \STATE $I^{'} = (Q^{'}, C^{'}, Ans^{'})$
        \STATE $\mathcal{D}^{'}.\text{append}(I^{'})$
    \ENDWHILE

    \STATE $\mathcal{D}_{\text{new}} = \text{RandomShuffle}(\mathcal{D}^{'} + \mathcal{D}[\alpha N :])$

    \red{// Training Stage-1: Refining}
   \FOR{iteration in Stage 1}
        \STATE Generate responses $o_i$
        \FOR{each response $o_i$}
            \STATE $\text{Thinking}_{\text{gen}}, \text{Ans}_{\text{gen}} = \text{Split}(o_i)$
            \STATE $R_{\text{structure}} = \bigcap_{i=1}^{M} \mathbb{I}_{\text{re.find}(\text{chr}(A + i), \text{Thinking}_{\text{gen}})}$
            \STATE $R = \alpha_1 R_{\text{format}} + \beta_1 R_{\text{structure}}$
            \STATE Compute advantage $\hat{A}_i$ and optimize $\mathcal{J}_{GRPO}(\theta)$
        \ENDFOR
   \ENDFOR

   \red{// Training Stage-2: Monitoring}
   \FOR{iteration in Stage 2}
        \STATE Generate responses $o_i$
        \FOR{each response $o_i$}
            \STATE $\text{Thinking}_{\text{gen}}, \text{Ans}_{\text{gen}} = \text{Split}(o_i)$
            \STATE $\text{Ans}_{\text{parsed}} = \text{LLM}_{\text{parse}}(\text{Thinking}_{\text{gen}})$
            \STATE $R_{\text{consistency}} = \mathbb{I}_{Ans_{\text{parsed}} = Ans_{\text{gen}}}$
            \STATE $R_{\text{ans}} = \mathbb{I}_{Ans_{\text{gen}} = Ans_{\text{ref}}}$
            \STATE $R = \gamma R_{\text{format}} + \alpha_2 R_{\text{consistency}} + \beta_2 R_{\text{ans}}$
            \STATE Compute advantage $\hat{A}_i$ and optimize $\mathcal{J}_{GRPO}(\theta)$
        \ENDFOR
   \ENDFOR
\end{algorithmic}
}
   \caption{\textsc{CLARity}}
   \label{alg:data}
\end{algorithm*}

\begin{CJK}{UTF8}{gbsn}
\begin{table*}[]
\scriptsize
\centering
\begin{tabular}{p{0.9\columnwidth}p{1.05\columnwidth}}
\toprule
\multicolumn{2}{p{2\columnwidth}}{\textbf{问题:} 甲公司借用乙公司的一套设备，在使用过程中不慎损坏一关键部件，于是甲公司提出买下该套设备，乙公司同意出售。双方还口头约定在甲公司支付价款前，乙公司保留该套设备的所有权。不料在支付价款前，甲公司生产车间失火，造成包括该套设备在内的车间所有财物被烧毁。对此，下列哪些选项是正确的?} \\
\multicolumn{2}{p{2\columnwidth}}{\texttt{Question:} On matters related to the Civil Procedure Law, the Administrative Law, the Administrative Litigation Law, and other relevant laws and regulations, which of the following statements is incorrect? } \\
\midrule
\multicolumn{2}{p{2\columnwidth}}{选项：
\textbf{A}: 乙公司已经履行了交付义务，风险责任应由甲公司负担 \quad
\textbf{B}: 在设备被烧毁时，所有权属于乙公司，风险责任应由乙公司承担 \quad
\textbf{C}: 设备虽然已经被烧毁，但甲公司仍然需要支付原定价款 \quad
\textbf{D}: 双方关于该套设备所有权保留的约定应采用书面形式} \\
\multicolumn{2}{p{2\columnwidth}}{\texttt{Options}:
\textbf{A}: Company B has fulfilled its delivery obligation, and the risk responsibility should be borne by Company A. \quad
\textbf{B}: When the equipment was destroyed, the ownership remained with Company B, and the risk responsibility should be borne by Company B. \quad
\textbf{C}: Although the equipment has been destroyed, Company A is still required to pay the original price. \quad
\textbf{D}: The agreement between both parties regarding the retention of ownership of the equipment should be in writing.} \\
\texttt{Answer}: \textbf{AC} \\
\midrule
\multicolumn{2}{c}{\textit{\textbf{Qwen2.5-7B-Instruct}}} \\
\midrule
<思考> 本题涉及所有权保留买卖合同中的风险责任分配问题。根据《合同法》第142条，标的物毁损、灭失的风险，在标的物交付之前由出卖人承担，交付之后由买受人承担，但法律另有规定的除外。本题中，甲公司与乙公司口头约定在甲公司支付价款前，乙公司保留该套设备的所有权，因此在甲公司支付价款前，设备的所有权仍属于乙公司。根据《合同法》第142条，设备被烧毁的风险应由乙公司承担。选项A错误，选项B正确。根据《合同法》第167条，当事人约定标的物所有权保留的，出卖人对标的物保留的所有权，未经登记，不得对抗善意第三人。本题中，甲公司与乙公司未采用书面形式，因此该约定不能对抗善意第三人。选项D正确。根据《合同法》第142条，设备被烧毁的风险应由乙公司承担，因此甲公司无需支付原定价款。选项C错误。</思考><回答> 本题答案为{B}和{D}。 </回答> & <Thought> This question concerns risk allocation in a retention of title sales contract. According to Article 142 of the Contract Law, the risk of damage or loss of the subject matter is borne by the seller before delivery and by the buyer after delivery, unless otherwise specified by law. Here, Company A and Company B orally agreed that Company B retains ownership of the equipment until Company A pays, so ownership remains with Company B until payment. According to Article 142, the risk of destruction is borne by Company B. Option A is incorrect, and option B is correct. According to Article 167 of the Contract Law, a retention of title agreement must be registered to be enforceable against bona fide third parties. Since the agreement was not in writing, it cannot be asserted against a bona fide third party. Therefore, option D is correct. As the risk of destruction is borne by Company B, Company A does not need to pay the original price. Option C is incorrect.</Thought>
<Answer> The correct answers are {B} and {D}. </Answer> \\
\midrule
\multicolumn{2}{c}{\textit{\textbf{Vanilla Reinforcement Learning}}} \\
\midrule
<思考> 根据《中华人民共和国合同法》的规定，标的物毁损、灭失的风险，在标的物交付之前由出卖人承担，交付之后由买受人承担，但法律另有规定的除外。在本案例中，甲公司和乙公司之间存在设备买卖合同，且双方口头约定在甲公司支付价款前，乙公司保留设备的所有权。因此，设备在交付给甲公司后，风险责任应由甲公司承担。但是，由于设备被烧毁，设备的所有权归属问题需要明确。根据《中华人民共和国合同法》第一百三十四条的规定，当事人可以在买卖合同中约定买受人未履行支付价款或者其他义务的，标的物的所有权属于出卖人。因此，双方关于设备所有权保留的约定是有效的。但是，由于设备被烧毁，设备的价值已经不存在，因此甲公司无需支付原定价款。

选项A：根据上述分析，设备在交付给甲公司后，风险责任应由甲公司承担，因此该选项正确。

选项B：由于设备在交付给甲公司后，风险责任应由甲公司承担，因此该选项错误。

\red{选项C：由于设备被烧毁，设备的价值已经不存在，因此甲公司无需支付原定价款，因此该选项错误。}

选项D：根据《中华人民共和国合同法》第一百三十四条的规定，当事人可以在买卖合同中约定买受人未履行支付价款或者其他义务的，标的物的所有权属于出卖人。因此，双方关于设备所有权保留的约定应采用书面形式，因此该选项正确。综上所述，正确的选项是A、\red{C}。</思考><回答> {A\red{C}} </回答> &  <Thought> According to the Contract Law of the People's Republic of China, the risk of damage or loss of the subject matter is borne by the seller before delivery and by the buyer after delivery, unless otherwise specified by law. In this case, there is a sales contract between Company A and Company B, with an oral agreement that Company B retains ownership of the equipment until Company A pays. Therefore, after delivery, the risk is borne by Company A. However, as the equipment was destroyed, the issue of ownership needs to be clarified. According to Article 134 of the Contract Law, the parties may agree that ownership remains with the seller if the buyer fails to pay the price or fulfill other obligations. Therefore, the retention of ownership agreement is valid. However, since the equipment is destroyed, its value no longer exists, and Company A does not need to pay the original price.

Option A: Based on the above analysis, after delivery, the risk responsibility is borne by Company A, so this option is correct.

Option B: Since the risk responsibility is borne by Company A after delivery, this option is incorrect.

\red{Option C: Since the equipment is destroyed and its value no longer exists, Company A does not need to pay the original price, so this option is incorrect.}

Option D: According to Article 134 of the Contract Law, the retention of ownership agreement must be in writing, so this option is correct. Therefore, the correct answers are A and \red{C}. </Thought>
<Answer> {A\red{C}} </Answer> \\
\midrule
\multicolumn{2}{c}{\textit{\textbf{W/o Stage-1}}} \\
\midrule
<思考> 该题涉及所有权保留买卖合同中的风险责任和价款支付问题。正确答案是A和C。A选项正确，因为设备已经交付，风险责任由买方承担。C选项正确，即使设备毁损，甲公司仍需支付原定价款。B选项错误，因为设备毁损时所有权属于乙公司，风险责任应由甲公司承担。D选项错误，关于所有权保留的约定可以采用口头形式。因此，答案是AC。 </思考><回答> {AC} </回答> & <Thought> This question involves risk responsibility and payment issues in a retention of title sales contract. The correct answers are A and C. Option A is correct because the equipment has been delivered, and the risk is borne by the buyer. Option C is correct because Company A must still pay the original price, even if the equipment is damaged. Option B is incorrect because ownership was with Company B when the equipment was damaged, and the risk responsibility should be borne by Company A. Option D is incorrect because the retention of title agreement can be oral. Therefore, the answer is AC. </Thought>
<Answer> {AC} </Answer> \\
\midrule
\multicolumn{2}{c}{\textit{\textbf{CLARity}}} \\
\midrule
<思考> 本题考查所有权保留买卖中的风险承担问题。

根据《合同法》第一百四十一条规定，出卖人交付标的物后，风险责任一般由买受人承担。但是，本题中甲公司和乙公司有特别约定，在甲公司支付价款前，乙公司保留该套设备的所有权。因此，根据《合同法》第一百四十二条规定，标的物毁损、灭失的风险，在标的物交付之后由买受人承担，但法律另有规定或者当事人另有约定的除外。本题中，风险责任应由甲公司承担，因此选项A正确，选项B错误。

根据《合同法》第一百四十七条规定，出卖人交付标的物后，买受人在标的物毁损、灭失的风险发生后尚未支付价款的，出卖人有权要求买受人支付全部价款或者就标的物的毁损、灭失获得赔偿。因此，即使设备被烧毁，甲公司仍然需要支付原定价款，因此选项C正确。

根据《合同法》第一百六十七条规定，出卖人保留所有权的，应当与买受人约定所有权保留的期间。因此，关于所有权保留的约定应当采用书面形式，因此选项D错误。

因此，正确的选项是A和C。{AC} </思考><回答> {AC} </回答> & <Thought> This question examines the issue of risk allocation in a retention of title sales contract.

According to Article 141 of the Contract Law, after the seller delivers the subject matter, the risk is generally borne by the buyer. However, in this case, Company A and Company B have a special agreement that Company B retains ownership of the equipment until Company A pays. Therefore, according to Article 142 of the Contract Law, the risk of damage or loss is borne by the buyer after delivery, unless otherwise specified by law or agreed by the parties. In this case, the risk is borne by Company A, so Option A is correct, and Option B is incorrect.

According to Article 147 of the Contract Law, if the buyer has not paid the price after the risk of damage or loss occurs, the seller has the right to demand full payment or compensation for the damage or loss. Therefore, even if the equipment is destroyed, Company A must still pay the original price, so Option C is correct.

According to Article 167 of the Contract Law, if the seller retains ownership, the period of retention must be agreed upon with the buyer. Therefore, the retention of title agreement should be in writing, making Option D incorrect.

Thus, the correct answers are A and C. {AC} </Thought>
<Answer> {AC} </Answer> \\
\bottomrule
\end{tabular}
\caption{A model response example from different methods in Jec-QA validation set. \red{Red part} contains inconsistency.}
\label{tab:app:casestudy:jecqa}
\end{table*}
\end{CJK}

\begin{table*}
\centering
\scriptsize
\setlength{\tabcolsep}{5pt}
\begin{tabular}{p{2.0\columnwidth}r}
\toprule
\texttt{Question:} Question:  A 33-year-old man presents to the emergency department with joint pain. He states that since yesterday he has had joint pain, and today he noticed a rash. The patient does not have a significant past medical history. The patient smokes cigars, drinks 1 to 4 drinks per day, is sexually active, and uses cocaine occasionally. His temperature is 101°F (38.3°C), blood pressure is 125/65 mmHg, pulse is 80/min, respirations are 14/min, and oxygen saturation is 98\% on room air. Physical exam reveals pustular skin lesions and an inflamed and painful right knee that is warm to the touch. Which of the following is the most appropriate treatment for the most likely diagnosis? \\
\texttt{Answer}: \textbf{B} \\
\midrule
\texttt{Options:} \textbf{A}: Azithromycin \quad \textbf{B}: Cefixime and azithromycin \quad \textbf{C}: Ceftriaxone \quad \textbf{D}: Ceftriaxone and vancomycin \quad \textbf{E}: Piperacillin and tazobactam \\
\midrule
\multicolumn{1}{c}{\textbf{\textit{Qwen2.5-7B-Instruct}}} \\
\midrule
<think> To determine the most appropriate treatment, we need to consider the patient's symptoms and the likely diagnosis. The patient presents with joint pain and a rash, which could indicate a variety of conditions, but the presence of pustular skin lesions and an inflamed, warm, and painful knee suggests a possible bacterial infection, such as gonorrhea or chlamydia, which can present with these symptoms.

- **Azithromycin (A)**: This is an antibiotic commonly used for chlamydia and other bacterial infections. However, it may not be sufficient for a more severe infection like gonorrhea. 

- **Cefixime and azithromycin (B)**: Cefixime is a cephalosporin antibiotic that is effective against a wide range of bacteria, including those causing gonorrhea. Azithromycin is also included to cover for chlamydia. This combination is often used for treating sexually transmitted infections (STIs).

- **Ceftriaxone (C)**: This is a broad-spectrum cephalosporin that is highly effective against gonorrhea. It is often used as a single-dose treatment for gonorrhea.

- **Ceftriaxone and vancomycin (D)**: Ceftriaxone is effective against gonorrhea, but adding vancomycin is typically not necessary unless there is a concern for resistant strains or other complications.

- **Piperacillin and tazobactam (E)**: This combination is used for more severe infections, such as those caused by gram-negative bacteria, but it is not the first-line treatment for the likely diagnosis in this case. Given the symptoms and the likely diagnosis of gonorrhea, the most appropriate treatment would be a combination that effectively covers both the likely bacterial cause and any potential co-infections. </think><answer> The most likely diagnosis is gonorrhea, given the symptoms and the patient's sexual activity. The most appropriate treatment would be a combination of Ceftriaxone and Azithromycin (B) to cover both the bacterial infection and the potential for co-infection with chlamydia. </answer>B \\
\midrule
\multicolumn{1}{c}{\textbf{\textit{Vanilla Reinforcement Learning}}} \\
\midrule
<think> To determine the most likely diagnosis and appropriate treatment, let's analyze the symptoms and physical exam findings. The patient has joint pain, a rash, and pustular skin lesions, which are consistent with gonococcal arthritis and dermatitis (gonococcal infection). Gonococcal infection is caused by Neisseria gonorrhoeae and can present with these symptoms.
The most likely diagnosis is gonococcal arthritis and dermatitis. The treatment for gonococcal infection is usually ceftriaxone, which is a broad-spectrum antibiotic effective against Neisseria gonorrhoeae.
Let's consider the treatment options: 

A: Azithromycin - This is an antibiotic used for chlamydia and other infections, but not for gonorrhea.

B: Cefixime and azithromycin - Cefixime is an oral cephalosporin, but it is not as effective as ceftriaxone for gonorrhea. 

\red{C: Ceftriaxone - This is the standard treatment for gonorrhea.}

\red{D: Ceftriaxone and vancomycin - Vancomycin is used for methicillin-resistant Staphylococcus aureus (MRSA) and other Gram-positive infections, but it is not necessary for gonorrhea.}

E: Piperacillin and tazobactam - This is a broad-spectrum antibiotic combination used for Gram-negative and Gram-positive infections, but it is not the standard treatment for gonorrhea. </think><answer>{D}</answer> \\
\midrule
\multicolumn{1}{c}{\textbf{\textit{W/o Stage-1}}} \\
\midrule
<think> To determine the most appropriate treatment, we need to first identify the most likely diagnosis based on the patient's symptoms. The patient presents with pustular skin lesions and a painful, warm knee, which are consistent with gonococcal arthritis or skin infection caused by Neisseria gonorrhoeae or Chlamydia trachomatis. However, the presence of pustular lesions is more indicative of a bacterial skin infection, likely caused by Staphylococcus aureus, particularly methicillin-resistant Staphylococcus aureus (MRSA).

Given the patient's history of cocaine use and occasional sexual activity, we need to consider the possibility of sexually transmitted infections (STIs) such as gonorrhea or chlamydia. However, the pustular lesions are more suggestive of a skin infection rather than a STI.

The most likely diagnosis is a skin infection caused by MRSA, which is common in individuals with risk factors for skin infections, such as cocaine use and smoking.

The treatment for MRSA skin infections typically includes antibiotics effective against both methicillin-susceptible and methicillin-resistant Staphylococcus aureus. The most appropriate antibiotic for MRSA is usually a beta-lactam antibiotic with beta-lactamase inhibition, such as cloxacillin, or a glycopeptide like vancomycin. However, for outpatient treatment, oral antibiotics like clindamycin or doxycycline are often used.

Given the options, the most appropriate treatment for a likely MRSA skin infection is:

A: Azithromycin - This is typically used for chlamydia and mycoplasma infections, not for MRSA.

B: Cefixime and azithromycin - Cefixime is not effective against MRSA, and azithromycin is not the first-line treatment for MRSA.

C: Ceftriaxone - This is typically used for gonorrhea and chlamydia, not for MRSA.

D: Ceftriaxone and vancomycin - Ceftriaxone is not effective against MRSA, and vancomycin is used for severe cases or when MRSA is suspected.

E: Piperacillin and tazobactam - This is typically used for Gram-negative infections and is not the first-line treatment for MRSA. The most appropriate treatment for a likely MRSA skin infection is a beta-lactam antibiotic with beta-lactamase inhibition, such as dicloxacillin or a glycopeptide like vancomycin. However, among the options provided, the most appropriate treatment would be a combination of an antibiotic effective against MRSA and a broad-spectrum antibiotic to cover other potential infections.

Given the options, the most appropriate treatment is Ceftriaxone and vancomycin, which is typically used for severe cases or when MRSA is suspected. Therefore, the most appropriate treatment for the most likely diagnosis is D: Ceftriaxone and vancomycin.</think><answer>{D}</answer> \\
\midrule
\multicolumn{1}{c}{\textbf{\textit{CLARity}}} \\
\midrule
<think>To determine the most appropriate treatment, we first need to identify the most likely diagnosis based on the patient's symptoms and physical exam findings.

The patient presents with fever, joint pain, and pustular skin lesions. These symptoms are characteristic of gonococcal arthritis, which is caused by Neisseria gonorrhoeae. Gonococcal arthritis typically affects large joints and is associated with other signs of gonococcal infection such as skin lesions.

The most common approach to treating gonococcal arthritis is to treat the underlying gonorrhea infection. The standard treatment for uncomplicated gonorrhea includes a single intramuscular dose of ceftriaxone. Given that the patient also has skin lesions, which could indicate disseminated gonococcal infection (DGI), a more aggressive treatment regimen is recommended.

The Centers for Disease Control and Prevention (CDC) recommends ceftriaxone 250 mg IM plus either azithromycin 1 g orally or cefixime 400 mg orally for the treatment of DGI. Therefore, the most appropriate treatment for this patient would be ceftriaxone and either azithromycin or cefixime.
Let's evaluate each option:

A: Azithromycin - This is a suitable oral antibiotic for co-treatment with ceftriaxone in DGI.

B: Cefixime and azithromycin - This is the recommended combination according to CDC guidelines.

C: Ceftriaxone - This is part of the treatment but needs to be combined with an oral antibiotic.

D: Ceftriaxone and vancomycin - Vancomycin is not typically used for gonorrhea unless there is evidence of penicillin allergy or resistance.

E: Piperacillin and tazobactam - This combination is used for more severe infections like endocarditis but is not the standard treatment for gonococcal arthritis. 
Given the guidelines, the most appropriate treatment is Ceftriaxone 250 mg IM plus Azithromycin 1 g orally. Therefore, the correct answer is B.</think><answer>{B}</answer> \\
\bottomrule
\end{tabular}
\caption{A model response example from different methods in MedQA validation set. \red{Red part} contains inconsistency.}
\label{tab:app:casestudy:medqa}
\end{table*}

\end{document}